%% file: paper.tex
\documentclass[ba]{imsart}

\RequirePackage{amsthm,amsmath,amsfonts,amssymb}
\RequirePackage[authoryear]{natbib}
\RequirePackage{chapterbib}
\RequirePackage[colorlinks,citecolor=blue,urlcolor=blue,backref=page,backref=page]{hyperref}
\RequirePackage{graphicx}
\RequirePackage{multirow}

\pubyear{2026}
\arxiv{2010.00000}
\volume{TBA}
\issue{TBA}
\firstpage{1}
\lastpage{1}

\startlocaldefs
\theoremstyle{plain}

\newtheorem{theorem}{Theorem}[section]
\newtheorem{lemma}[theorem]{Lemma}
\newtheorem{corollary}[theorem]{Corollary}
\newtheorem{proposition}[theorem]{Proposition}

\theoremstyle{definition}

\theoremstyle{remark}

\newcommand{\plus}{\raisebox{.4\height}{\scalebox{.6}{+}}}
\newcommand{\mref}[1]{(\ref{#1})}
\newcommand{\Prob}{\operatorname{P}}
\newcommand{\given}{\mid}
\newcommand{\X}{\mathbf{X}}
\newcommand{\G}{\mathcal{G}}
\newcommand{\D}{\mathcal{D}}
\newcommand{\PXi}{\Pi_{X_i}}
\newcommand{\XPi}{X_i \given \PXi}
\newcommand{\KL}{\mathsf{KL}}
\newcommand{\SHD}{\mathsf{SHD}}
\newcommand{\SID}{\mathsf{SID}}
\newcommand{\wSHD}{\widetilde{\mathsf{SHD}}}
\newcommand{\wSID}{\widetilde{\mathsf{SID}}}
\newcommand{\PD}{\mathsf{PD}}
\newcommand{\ind}{{{\mathchoice {\rm 1\mskip-4mu l} {\rm 1\mskip-4mu l}
            {\rm 1\mskip-4.5mu l} {\rm 1\mskip-5mu l}}}}

\DeclareMathOperator{\rarr}{\rightarrow}
\DeclareMathOperator{\larr}{\leftarrow}

\endlocaldefs

\begin{document}

\begin{frontmatter}
\title{A Causal Energy Prior for Learning Directed Acyclic Graphs}
\runtitle{A Causal Energy Prior for Learning DAGs}

\begin{aug}
\author[A]{%
  \fnms{Marco}~\snm{Scutari}\ead[label=ms]{scutari@bnlearn.com}%
  \orcid{0000-0002-2151-7266}%
},
\author[B]{%
  \fnms{Noriaki}~\snm{Sato}\ead[label=ns]{noriakis@ims.u-tokyo.ac.jp}%
  \orcid{0000-0001-7721-9359}%
},
\author[C]{%
  \fnms{Shuichi}~\snm{Kawano}\ead[label=sk]{skawano@math.kyushu-u.ac.jp}%
  \orcid{0000-0002-0804-0141}%
},
\author[D]{%
  \fnms{Rui}~\snm{Yamaguchi}\ead[label=ry]{r.yamaguchi@aichi-cc.jp}%
  \orcid{0000-0003-1224-227X}%
}
\and
\author[B]{%
  \fnms{Seiya}~\snm{Imoto}\ead[label=si]{imoto@hgc.jp}%
  \orcid{0000-0002-2989-308X}%
}

\address[A]{Istituto Dalle Molle di Studi sull'Intelligenza Artificiale (IDSIA),
  Lugano, Switzerland\printead[presep={,\ }]{ms}}
\address[B]{The Institute of Medical Science, The University of Tokyo, Tokyo,
  Japan\printead[presep={,\ }]{ns,si}}
\address[C]{Faculty of Mathematics, Kyushu University, Fukuoka,
  Japan\printead[presep={,\ }]{sk}}
\address[D]{Division of Cancer System Biology, Aichi Cancer Center Research
  Institute; Division of Cancer Informatics, Nagoya University Graduate School
  of Medicine, Nagoya, Japan\printead[presep={,\ }]{ry}}
\runauthor{M. Scutari et al.}
\end{aug}

\begin{abstract}
  Bayesian learning of directed acyclic graphs (DAGs) has long relied on non- or
  weakly informative priors, on the assumption that the information needed for
  an informative one is rarely available. In genetics and systems biology,
  annotation databases now make it possible to assemble a prior DAG before
  learning from data. Existing informative priors are local: they score a DAG
  arc by arc and cannot encode how interventions propagate along its paths.
  Thus, we propose the local-global energy prior (LGE), a Gibbs distribution
  over DAGs whose energy combines a local divergence from the prior DAG, the
  Structural Hamming Distance, and a causal one, the Structural Intervention
  Distance. We derive rules to calibrate its decay and the weight of the two
  divergences from the prior DAG alone. We quantify how much prior mass stays on
  the true DAG, and how much data overturn the prior, when the prior DAG is
  partially incorrect. In simulations, LGE improves both the structural accuracy
  of maximum a posteriori DAGs and the fit of the corresponding Bayesian
  networks to the true distribution over non- and weakly informative priors,
  even when half of the arcs of the prior DAG are wrong. We also showcase LGE on
  a gene regulatory network learned from single-cell kidney data.
\end{abstract}

\begin{keyword}[class=MSC2020]
\kwd[Primary ]{62H22}
\kwd[; secondary ]{62F15}
\kwd{62D20}
\end{keyword}

\begin{keyword}
\kwd{Directed Acyclic Graphs}
\kwd{Energy-Based Priors}
\kwd{Structure Learning}
\kwd{Prior Calibration}
\kwd{Structural Intervention Distance}
\end{keyword}

\end{frontmatter}

\section{Introduction}
\label{sec:introduction}

Directed acyclic graphs (DAGs) are an integral part of foundational models in
statistics and machine learning. In \emph{Bayesian networks}
\citep[BNs;][]{lauritzen}, they encode patterns of conditional independence
between the variables of a multivariate distribution via graphical separation;
in \emph{causal networks} \citep{causality}, they encode the cause-effect
relationships between the parts of a data generating process. Both have been
widely applied in the life and clinical sciences, from sequencing
\citep{carapito} and molecular data \citep{omics} to epidemiology
\citep{epidemiology}, electronic health records \citep{pfizer24} and clinical
conditions \citep{jms25,aneurysm22,lim22,belyaeva}.

This interest has in turn motivated progress in posterior DAG learning for BNs:
\emph{maximum a posteriori} (MAP) search has improved in accuracy and
scalability \citep{xges,suzuki}, marginal likelihoods are available for
increasingly complex data \citep{abn,ijar21-laura}, and Markov chain Monte Carlo
(MCMC) samplers over DAGs \citep{giusi-mcmc1,giusi-uai,marco-hybrid} have
evolved to explore the enormous search space \citep{harary}.

In contrast, graphical prior specification has been less explored. Classical BN
literature assumes a non-informative, uniform prior over the space of DAGs for
both computational and mathematical simplicity \citep{friedman2}. Later
proposals switched to weakly informative priors that control sparsity either by
modelling node in-degrees \citep{gprior} or by setting a uniform prior on the
inclusion and direction of individual arcs \citep{pgm16}. These works
assume that specifying granular, informative priors over DAGs is unfeasible
because of the intrinsic limits of eliciting priors from experts and formalising
them mathematically. When otherwise feasible, \citet{csprior} defined prior
inclusion probabilities for a subset of relevant arcs, while \citet{mukherjee}
proposed a systematic way to put priors on a small set of more complex graphical
features using energy-based priors.

This assumption is now often incorrect in the life and clinical
sciences. Efforts to catalogue the experimental results available in the
literature have produced comprehensive \emph{annotation databases} for several
types of molecular data: KEGG \citep{kegg}, DoRothEA \citep{dorothea}, the GWAS
Catalog \citep{catalog} and 100$\plus$ others are discussed in
\citet{nar-collection}. Their scope and level of detail vary, but all contain
the information we need to link medical conditions and the different molecules
involved in complex biological processes, within and across molecule types.
These links can form an \emph{a priori} DAG which we can use to build an
informative graphical prior distribution; for instance, \citet{isci} combined
pathway, protein interaction and co-expression databases into a prior for
learning gene interaction networks. Literature mining with large language models
\citep{nlp1,nlp2} can tentatively fill the gaps of such databases, if less
reliably.

Estimating the effects of interventions is a key use of both BNs and causal
networks: in systems biology, network models rank candidate treatment targets,
and only a model that reflects the effects of pharmacological interventions can
do so reliably \citep{carapito,klinke}. In clinical studies, separating the
direct and indirect effects of an intervention requires an accurate map of the
causal pathways involved \citep{uhler}. Those effects spread along the paths in
the DAG, so two DAGs that differ by a single arc may agree on every causal
effect or disagree on many of them. A prior defined over individual arcs cannot
tell which is the case, regardless of how much prior information is available.

Structural divergences give formal ways to measure how two DAGs differ, and
have progressively moved from local to causal criteria. Older divergences from
the BN literature, such as the Structural Hamming Distance \citep[SHD;][]{mmhc},
are local: they compare two DAGs at the level of individual arcs, regardless of
where those arcs appear. \emph{Causal divergences} such as the Structural
Intervention Distance \citep[SID;][]{peters} instead count the pairs of nodes
for which two DAGs disagree on how the effects of interventions propagate along
the paths connecting them. However, they have not been used to inform DAG
learning: the causal discovery literature relies on identifiability through
distributional assumptions and maximum likelihood estimation rather than on
Bayesian learning. To our knowledge, the prior we propose is the first to
incorporate a causal divergence and to encode prior information on the
interventional behaviour of a DAG.

We call it the \emph{local-global energy} (LGE) prior, after the energy-based
prior over DAGs from \citet{imoto04}: it anchors Bayesian learning to a prior
DAG, which we assume to be known but not necessarily correct, through a local
(SHD) and a global (SID) structural divergence. Our contributions are the prior
itself (Section~\ref{sec:lge}); rules to calibrate its two hyperparameters from
the prior DAG alone (Sections~\ref{sec:calibration} and~\ref{sec:weight});
bounds on the prior mass it leaves on the true DAG when the prior DAG is wrong,
and on the data needed to overturn it (Section~\ref{sec:robustness}); a
simulation study of MAP DAG learning (Section~\ref{sec:sims}); and an
application to kidney gene expression data (Section~\ref{sec:kidney}).
Section~\ref{sec:notation} introduces the background and notation, and
Section~\ref{sec:conclusions} discusses LGE's potential and limitations. Proofs
as well as additional technical material and results are provided in the
Supplementary Material.

\section{Background and Notation}
\label{sec:notation}

This section introduces the notation for DAGs, BNs and causal networks. We also
review the graphical priors used in BN structure learning and the two structural
divergences, SHD and SID, from which we build the LGE prior in
Section~\ref{sec:lge}.

Let $\G = (V, A)$ be a DAG with arc set $A$ and $N$ nodes
$V = \{ v_i, i = 1, \ldots, N \}$. The arc set
$A = \{ (v_i, v_j) : v_i \rarr v_j, i \neq j \}$ defines which nodes are
connected by an arc and the arc's direction, and is often represented as an
adjacency matrix. Furthermore, there are $P = {N \choose 2}$ unordered
\emph{node pairs} $\{ v_i, v_j \}, i \neq j$, each in one of three states:
$(v_i, v_j) \in A$, $(v_j, v_i) \in A$ or $(v_i, v_j) \not\in A \wedge
(v_j, v_i) \not\in A$. For two DAGs over the same nodes, we write
$\G' \subseteq \G''$ when $A' \subseteq A''$, that is, when $\G''$ contains
every arc of $\G'$.

In BNs, each node $v_i \in V$ is associated with a random variable $X_i \in \X$,
and $\G$ represents the conditional independence relationships between the
variables in $\X$. This construction leads to the decomposition
\begin{align}
  &\Prob(\X; \Theta) = \prod_{i = 1}^N \Prob(\XPi; \Theta_i),&
  &\Theta = \bigcup_{i = 1}^N \Theta_i
\label{eq:parents}
\end{align}
where $\PXi$ are the parents of $X_i$ in $\G$ and $\Theta_i$ its parameters, by
way of the Markov property \citep{lauritzen}. Different DAGs can encode the same
distribution: DAGs are grouped into equivalence classes, each represented
graphically by a completed partially directed acyclic graph in which only the
arcs that have the same direction in all the DAGs in the class are directed
\citep{chickering}. Thus, two DAGs $\G'$ and $\G''$ are compared using
the Structural Hamming Distance \citep[SHD;][]{mmhc}, which counts the number
of arcs differing between the completed partially directed acyclic graphs of
the respective equivalence classes. SHD is therefore a \emph{local} divergence.

In Bayesian structure learning, we estimate $\G$ from a sample $\D$ of $n$
observations from $\X$ through its posterior probability
\begin{equation*}
  \Prob(\G \given \D) \propto \Prob(\G)\Prob(\D \given \G),
\end{equation*}
where the marginal likelihood $\Prob(\D \given \G)$, like the other network
scores commonly used in its place, is decomposable: it is a product of local
scores, one for each node, each depending only on $X_i$ and its parents
\citep{friedman2}. The DAGs in an equivalence class have the same likelihood by
definition, so only an informative graphical prior $\Prob(\G)$ over the space of
DAGs with node set $V$ can result in them having different posterior
probabilities. The (non-informative) uniform prior $\Prob(\G) \propto 1$
\citep{ba12} cannot do this, since it assigns the same probability to all DAGs.
The weakly-informative marginal uniform prior \citep{pgm16} assigns probability
$p$ to each of the two possible arcs between each node pair, independently of
the other pairs: $\Prob(\G)$ then depends only on $|A|$, which is the same for
all the DAGs in an equivalence class. The same holds for the weakly-informative
priors $\Prob(\G) \propto \prod_i f(d_i)$ from \citet{gprior}, where $f$ weights
the in-degrees $d_i$: the arc reversals that connect the DAGs in an equivalence
class \citep{chickering} swap the in-degrees of their endpoints. Both priors are
invariant to relabelling the nodes, so they cannot encode any knowledge about
specific arcs; they can only control how many arcs a DAG has and how they are
spread over its nodes, which is why we call them ``weakly informative''.

The general informative prior from \citet{csprior} is constructed similarly to
the marginal uniform prior but places separate probabilities on the three
states of each node pair $\{v_i, v_j\}$ for which prior information is
available. The states of the remaining pairs are modelled with a marginal
uniform distribution. As \citet{cussens} note, the resulting prior is a maximum
entropy distribution over DAGs; energy priors make this construction explicit.
Let
$E(\G) \geqslant 0$ be an \emph{energy} that measures how much $\G$ disagrees
with the prior knowledge. Among the distributions over the DAGs with node set
$V$ that have a given expected energy, the one with the largest entropy,
which encodes no information beyond this constraint
\citep[Chapter~3]{robert07}, is the Gibbs distribution
\begin{equation*}
  \Prob(\G) \propto \exp\left\{-\beta E(\G)\right\}, \qquad \beta \geqslant 0,
\end{equation*}
in which DAGs with lower energy have higher prior probability. The parameter
$\beta$ is determined by that expected energy: $\beta = 0$ gives the uniform
prior, the expected energy decreases as $\beta$ grows, and the prior
concentrates on the DAGs with the smallest energy as $\beta \to \infty$. If
$E(\G)$ is a sum of terms over the node pairs, $\Prob(\G)$ is a product over
them, as the prior of \citet{csprior} is. The prior from \citet{imoto04}
organises the prior knowledge on the arcs $(v_i, v_j)$ into a matrix $U =
[u_{ij}]$ and takes $E(\G) = \sum_{(v_i, v_j) \in A} u_{ij}$.
Each $u_{ij}$ takes a small value if the prior knowledge supports $v_i \rarr
v_j$ and a larger value otherwise, so that the DAGs containing the supported
arcs have lower energy. \citet{imoto04} choose the two values by minimising the
same network score they learn the DAGs with, which requires bounding the
normalising constant of $\Prob(\G)$, intractable for all but the smallest DAGs.
This construction has been expanded to mixtures of priors
\citep{husmeier} and complex graphical features \citep{mukherjee}, and applied
to molecular data by \citet{isci}. The energy functions they used are either
sums over node pairs or functions of local features such as motifs and
degrees: none involves the paths in $\G$.

In contrast, causal discovery aims at learning DAGs that represent the
cause-effect relationships in the data generating process that $\D$ arises from
\citep{causality}. A causal network is a BN whose arcs are interpreted as direct
causes. An \emph{intervention} on $X_i$ fixes its value from outside the system:
it removes the arcs pointing to $v_i$ and leaves all other local distributions
unchanged. The \emph{causal effect} of $X_i$ on $X_j$ is the distribution of
$X_j$ under the intervention, as a function of the value given to $X_i$.
Interventions propagate only along the directed paths that leave $v_i$: $X_i$
can have an effect on $X_j$ only if $v_j$ is a descendant of $v_i$.
Observational data contain no interventions, but the effect can be estimated
from them by adjusting for a set of variables that contains no descendants of
the nodes that follow $v_i$ on the directed paths to $v_j$ and \emph{blocks}
every other path between $v_i$ and $v_j$, in the sense that the path no longer
transmits dependence once we condition on the set \citep{causality}. Such a set
is a \emph{valid adjustment set} for $(\G, v_i, v_j)$; the parents of $v_i$ are
always one. Therefore, DAGs are no longer naturally
grouped into equivalence classes: the DAGs in an equivalence class represent
the same probability distribution, and thus have the same likelihood, but they
represent different cause-effect relationships. In $v_i \rarr v_j$, intervening
on $X_i$ changes $X_j$; in $v_j \rarr v_i$, it does not. Thus, the two DAGs are
equivalent as BNs but not as causal networks.

\begin{figure}[t!]
  \includegraphics[width=\linewidth]{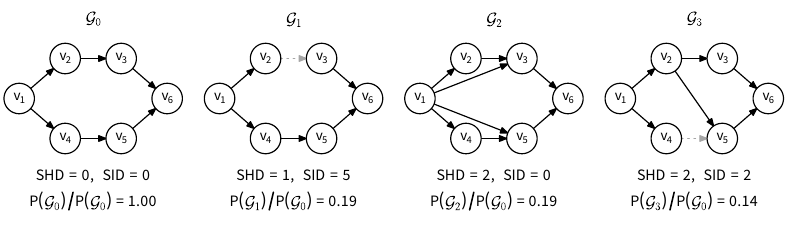}
  \caption{The LGE prior with $\beta = 10$ and $w = 0.5$ for a prior DAG $\G_0$
    and for three alternatives $\G_1$, $\G_2$, $\G_3$; dropped arcs are dotted
    grey lines.}
  \label{fig:example}
\end{figure}

As a result, causal networks are compared differently from BNs. In the causal
discovery literature, SHD compares two DAGs $\G'$ and $\G''$ directly, rather
than their equivalence classes \citep[see, for instance,][]{peters}. It becomes
a straightforward implementation of a Hamming distance between graphs, counting
the node pairs that are connected differently in $\G'$ and $\G''$ (an arc
between them appears in only one of the two DAGs, or appears in both with
opposite directions). Formally,\footnote{In all the divergences we use, the
first argument is the DAG under evaluation and the second is the reference it
is compared against. The distinction is immaterial for SHD, but it is not for
SID, for which \citet{peters} order the arguments the other way round.}
\begin{equation*}
  \SHD(\G', \G'') = \sum_{\{v_i, v_j\} \,:\, i < j} \ind\biggl(
    \{v_i, v_j\} \text{ is in different states in $\G'$ and $\G''$} \biggr),
\end{equation*}
where the three states of a node pair are those defined above. Each pair
contributes at most $1$, so a reversed arc counts as a single difference. This
is the definition we use throughout the rest of the paper. Causal networks can
also be compared in terms of how differently interventions propagate through
them. Let
\begin{equation*}
  R(\G) = \bigl\{ (v_i, v_j) : i \neq j, \text{$\G$ contains a directed path
    from $v_i$ to $v_j$} \bigr\}
\end{equation*}
be the \emph{interventional reachability set} of $\G$, that is, the pairs in
which $v_j$ is a descendant of $v_i$, so that intervening on $X_i$ can alter
$X_j$. The first such divergence was the Structural Intervention Distance
\citep{peters},
\begin{equation*}
  \SID(\G', \G'') = \sum_{(v_i, v_j) \,:\, i \neq j} \ind\biggl(
    \Pi_{v_i}(\G') \text{ is not a valid adjustment set for }
    (\G'', v_i, v_j) \biggr).
\end{equation*}
SID counts the pairs $(v_i, v_j)$ for which the parent set of $v_i$ in the DAG
under evaluation $\G'$ is not a valid adjustment set in the reference $\G''$,
either because the two DAGs disagree on whether $v_j$ is reachable from $v_i$,
or because that parent set fails to block one of the other paths between them.
Formally, $\Pi_{v_i}(\G')$ is a valid adjustment set for $(\G'', v_i, v_j)$ if
either $v_j \in \Pi_{v_i}(\G')$ and $(v_i, v_j) \not\in R(\G'')$; or $v_j
\not\in \Pi_{v_i}(\G')$, no node in $\Pi_{v_i}(\G')$ is a descendant of any node
other than $v_i$ lying on a directed path from $v_i$ to $v_j$ in $\G''$, and
$\Pi_{v_i}(\G')$ blocks all paths between them other than the directed ones.
Unlike SHD, SID is not symmetric in its arguments, and it is global with respect
to $\G''$: altering a single arc of $\G''$ can affect the paths between other
pairs of nodes. For instance, let $\G'$ and $\G''$ be the DAGs labelled $\G_1$
and $\G_0$ in Figure~\ref{fig:example}: $\G'$ differs from $\G''$ only in
missing $v_2 \rarr v_3$, so $\SHD(\G', \G'') = 1$; but $v_3$ has no parents in
$\G'$, and its empty parent set leaves open the paths through $v_2$ that connect
it to every other node in $\G''$, so $\SID(\G', \G'') = 5$. Despite its appeal,
SID's parent adjustment is only one of many ways of identifying a causal effect,
motivating several alternative proposals such as the adjustment identification
distances of \citet{henckel}, which use other adjustment sets (such as the
ancestors of $v_i$ or the optimal adjustment set).

Causal discovery does not use graphical priors: it obtains identifiability from
distributional assumptions \citep{lingam,anm,xu22,park,zigdag}, and uses the
divergences above to evaluate the DAGs it learns rather than to inform the
learning process.

\section{Proposal: The Local-Global Energy Prior (LGE)}
\label{sec:lge}

The LGE prior measures how far a candidate DAG $\G$ is from a non-empty
\emph{prior DAG} $\G_0 = (V, A_0)$, which encodes the available prior knowledge:
the farther $\G$ is from $\G_0$, the smaller its probability. It has two
hyperparameters, which we refer to throughout the paper: the \emph{weight} $w$,
which balances a local and a global measure of the distance between $\G$ and
$\G_0$; and the \emph{decay} $\beta$, which controls how quickly the prior
probability decreases as that distance grows. We define the prior below, give
rules to choose the decay in Section~\ref{sec:calibration} and the weight in
Section~\ref{sec:weight}, and study how they should change when $\G_0$ contains
errors in Section~\ref{sec:robustness}.

Consider the energy function
\begin{align}
  &E(\G, \G_0; w) = w \wSHD(\G, \G_0) + (1 - w) \wSID(\G, \G_0),&
  &w \in [0, 1]
\label{eq:energy}
\end{align}
and the associated LGE graphical prior
\begin{align}
  &\Prob(\G; \G_0, \beta, w) \propto
    \exp\left\{ -\beta E(\G, \G_0; w) \right\},&
  &\beta \geqslant 0.
\label{eq:lge}
\end{align}
The energy function in \mref{eq:energy} is a convex combination of a
\emph{local} graphical divergence, SHD, and a \emph{global} graphical
divergence, SID, between the candidate DAG $\G$ and the prior DAG $\G_0$. Both
appear rescaled, as $\wSHD(\G, \G_0) = \SHD(\G, \G_0) / |A_0|$ and $\wSID(\G,
\G_0) = \SID(\G, \G_0) / \{N(N - 1)\}$, to align their ranges; we explain this
choice in Appendix~\ref{app:rescaling:choice} of the Supplementary Material,
and check it empirically in Appendix~\ref{app:rescaling:sims}. Like the energy
priors in Section~\ref{sec:notation}, LGE is therefore the maximum entropy
prior over DAGs for a given expected divergence from $\G_0$, measured by errors
that are critical in practical use: SHD counts the arcs of $\G$ that are wrong
or missing, SID the causal effects that $\G$ would estimate with an invalid
adjustment set. We do not set that expected divergence directly: it is implied
by $\beta$, which we choose in Section~\ref{sec:calibration}.

LGE includes the informative priors of \citet{imoto04} and \citet{csprior} from
Section~\ref{sec:notation} as special cases with $w = 1$. Consider each node
pair $\{v_i, v_j\}$ in turn. If $v_i \rarr v_j$ is in $\G_0$, $\{v_i, v_j\}$
contributes $1$ to $\SHD(\G, \G_0)$ unless $v_i \rarr v_j$ is also in $\G$; if
$v_i$ and $v_j$ are not adjacent in $\G_0$, it contributes $1$ if $\G$ contains
an arc between them. Summing over the node pairs,
\begin{align*}
  &\SHD(\G, \G_0) = |A_0| + \sum_{(v_i, v_j) \in A} u_{ij},&
  &u_{ij} = \begin{cases}
    -1 & \text{if $(v_i, v_j) \in A_0$,} \\
    0  & \text{if $(v_j, v_i) \in A_0$,} \\
    1  & \text{otherwise.}
  \end{cases}
\end{align*}
The term $|A_0|$ does not depend on $\G$, so LGE with $w = 1$ is the energy
prior of \citet{imoto04} with this matrix $U = [u_{ij}]$ and decay $\beta /
|A_0|$. Similarly, the prior of \citet{csprior} that assigns probability $\pi$
to the state each node pair takes in $\G_0$ and $(1 - \pi)/2$ to each of the
other two states is
\begin{equation*}
  \Prob(\G) \propto \pi^{P - \SHD(\G, \G_0)} \{(1 - \pi)/2\}^{\SHD(\G,\G_0)},
\end{equation*}
that is, LGE with $w = 1$ and $\beta = |A_0| \log\{2\pi / (1 - \pi)\}$ for
$\pi \in [1/3, 1)$, with $\pi = 1/3$ giving the uniform prior. For $w < 1$, the
energy also contains $\wSID$, which is not a sum involving individual arcs or
node pairs; therefore, LGE cannot be written as either prior.

The two terms in \mref{eq:lge} penalise deviations from $\G_0$ differently, as
illustrated in Figure~\ref{fig:example}, which reports $\SHD$ and $\SID$ from a
prior DAG $\G_0$ with $N = 6$ nodes and \mbox{$|A_0| = 6$} arcs; the rescaled
divergences in \mref{eq:energy} are then $\wSHD = \SHD / 6$ and
$\wSID = \SID / 30$. $\wSHD(\G, \G_0)$ vanishes only when $\G = \G_0$, and
counts every false positive and false negative with respect to $\G_0$.
$\wSID(\G, \G_0)$ vanishes whenever $\G_0 \subseteq \G$: it ignores spurious
arcs and penalises only the arcs that appear in $\G_0$ but not in $\G$, in
proportion to the interventional effects that become unidentifiable in $\G$ due
to their omission. The weight $w$ determines their relative contributions, and
the decay $\beta$ sets how quickly $\Prob(\G; \G_0, \beta, w)$ falls as $\G$
moves away from $\G_0$, from the uniform prior at $\beta = 0$ to a point mass on
$\G_0$ as $\beta \to \infty$. In Figure~\ref{fig:example}, $\G_2$ adds two arcs
to $\G_0$, so $\SHD = 2$, but $\SID = 0$; $\G_1$ drops a single arc, so
$\SHD = 1$, but the now empty parent set of $v_3$ fails to identify $5$ of the
$30$ possible effects, so $\SID = 5$. With $w = 0.5$, both have energy
$0.5 \cdot 2/6 = 0.5 \cdot 1/6 + 0.5 \cdot 5/30 = 1/6$, and thus the same prior
probability $e^{-10/6} \approx 0.19$ relative to $\G_0$ at $\beta = 10$.
$\G_3$ drops one arc of $\G_0$ and adds another, so $\SHD = 2$, and fails to
identify $2$ effects, so $\SID = 2$: its energy
$0.5 \cdot 2/6 + 0.5 \cdot 2/30 = 1/5$ is the largest of the three, and its
relative prior probability $e^{-2} \approx 0.14$ the smallest, because it is
penalised by both terms.

The prior in \mref{eq:lge} is only defined up to the normalising constant
\begin{align*}
  &Z(\beta, w) = \sum_{\G \in \mathbb{D}_N}
    \exp\left\{ -\beta E(\G, \G_0; w) \right\},&
  &\text{where $\mathbb{D}_N$ is the set of all DAGs over $V$.}
\end{align*}
Evaluating it is intractable, since $|\mathbb{D}_N|$ grows super-exponentially
in $N$ \citep{harary}. Fortunately, $Z(\beta, w)$ cancels out from the ratios
that drive both MCMC and MAP DAG learning, and from the calibration rules for
$\beta$ and $w$ in Sections~\ref{sec:calibration} and~\ref{sec:weight}. Only
the robustness results in Section~\ref{sec:robustness} and the absolute rule of
Appendix~\ref{app:absolute} need bounds on it, which we derive in
Appendix~\ref{app:partition:bounds} of the Supplementary Material by relaxing
the acyclicity constraint as in \citet{imoto04}.

\subsection{Calibrating the Decay}
\label{sec:calibration}

To choose a suitable value for the decay $\beta$, we quantify the mass of the
immediate neighbourhood of the prior DAG $\G_0$ relative to that of $\G_0$
itself. Smaller masses correspond to a faster decay (that is, the prior is more
concentrated around $\G_0$). We do that by considering $\Prob(\G) /
\Prob(\G_0)$, in which the normalising constant simplifies: we first count and
characterise the DAGs one arc edit away from $\G_0$, then bound their energy,
and finally derive a rule that sets $\beta$ from how much of the prior mass we
allow them to take. In the rest of this section, we take $w > 0$: at $w = 0$,
$\wSHD$ drops out of the energy and there is no decay to calibrate.

Let
\begin{equation*}
  S_k(\G_0) = \left\{ \G \in \mathbb{D}_N : \SHD(\G, \G_0) = k \right\}
\end{equation*}
be the shell of the DAGs that are exactly $k$ arc edits away from $\G_0$. Thus,
the \emph{first shell} $S_1(\G_0)$ collects the DAGs obtainable from $\G_0$ with
a single arc addition, deletion or reversal. Counting the additions that do not
create cycles (Lemma~\ref{theo:ballsize} in Appendix~\ref{app:calibration})
gives
\begin{equation}
  \left| S_1(\G_0) \right| =
    N(N - 1) - |R_0| + \#\{\text{acyclic reversals}\},
\label{eq:ballsize}
\end{equation}
which is at most $N(N - 1)$. Thus, $|S_1(\G_0)|$ is large when the
interventional reachability set $R_0 = R(\G_0)$ of Section~\ref{sec:notation} is
small, that is, when $\G_0$ has few directed paths. For the $\G_0$ in
Figure~\ref{fig:example},
$N(N - 1) = 30$, $|R_0| = 11$ and all $6$ arcs can be reversed without creating
cycles, so $|S_1(\G_0)| = 25$: $6$ deletions, $6$ reversals and $13$ additions.

Every $\G \in S_1(\G_0)$ has $\SHD(\G, \G_0) = 1$ by construction. Additions
have \mbox{$\SID(\G, \G_0) = 0$} because $\G_0 \subseteq \G$; deletions and
reversals have $\SID(\G, \G_0) \leqslant 2(N - 1)$ \citep{peters}. Hence,
$E(\G, \G_0; w) \in [ w/|A_0|, w/|A_0| + 2(1-w)/N ]$ within $S_1(\G_0)$. The
lower bound is attained by all the additions, which outnumber deletions and
reversals by $O(N^2)$ to $O(N)$ for sparse DAGs: $\wSHD$ is constant on the
shell and dominates the calibration.

In turn, this allows us to write
\begin{equation*}
  \frac{\Prob(S_1(\G_0))}{\Prob(\G_0)} =
    \sum_{\G \in S_1(\G_0)} \exp\left\{ -\beta E(\G, \G_0; w) \right\}
    \leqslant \left| S_1(\G_0) \right|
      \exp\left\{ -\beta w / |A_0| \right\},
\end{equation*}
without depending on $Z(\beta, w)$. For a given \emph{mass ratio} $r > 0$, we
then choose $\beta$ so that the first shell carries at most $r$ times the prior
probability of $\G_0$ (with $r = 1$, $\G_0$ is at least as likely as all the
DAGs one arc edit away from it together):
\begin{align}
  &\beta \geqslant \frac{|A_0|}{w} \log \frac{\left| S_1(\G_0) \right|}{r}&
  &\Longrightarrow&
  &\Prob(S_1(\G_0)) \leqslant r \Prob(\G_0).
\label{eq:ballrule}
\end{align}
For comparison, letting a single DAG with one spurious arc have $r$ times the
prior probability of $\G_0$ only requires $\beta \geqslant (|A_0| / w)
\log(1/r)$: the shell rule adds $(|A_0| / w) \log |S_1(\G_0)|$ to this because
it controls the first shell as a whole.

We call \mref{eq:ballrule} the \emph{shell rule}. It constrains only the product
$\beta w$, which we call the \emph{local decay} because it is the decay applied
to $\wSHD$ in \mref{eq:energy}, and it scales with $|A_0|$. The rule looks only
at the first shell around $\G_0$, which is a vanishing fraction of the space of
DAGs: it contains at most $N(N - 1)$ DAGs, whereas the number of DAGs grows
super-exponentially in $N$. A natural question is then whether such a small
region allows for calibrating $\beta$ reliably, or whether the DAGs outside it
carry enough prior mass to be relevant. In Appendix~\ref{app:absolute} of the
Supplementary Material, we use the bounds on $Z(\beta, w)$ from
Appendix~\ref{app:partition:bounds} to derive a second rule that controls the
absolute probability $\Prob(\G_0)$, and hence the total mass of all the other
DAGs. The shell rule controls the first term of a series over the shells
$S_k(\G_0)$, $k = 1, 2, \ldots$, and the second rule its sum; including all the
shells beyond the first shifts $\beta$ by a fraction that vanishes as
$O(1 / \log N)$ as $N$ grows. Hence, the first shell largely determines
$\beta$. In addition, the shell rule does not depend on the bounds on
$Z(\beta, w)$, which are conservative, so we use it in practice and treat the
second rule as its justification.

As we will confirm in Section~\ref{sec:robustness}, what value of $\beta$ is
preferable depends on how much we trust the prior DAG to be correct: the shell
rule \mref{eq:ballrule} converts that trust, in the form of the mass ratio $r$
allowed to the first shell, into a value of $\beta$. In
Figure~\ref{fig:example}, with $w = 0.5$ and $r = 1$, the shell rule requires
$\beta \geqslant 12 \log 25 \approx 38.6$: at the $\beta = 10$ used in the
figure, the first shell carries up to $11$ times the prior probability of
$\G_0$.

\subsection{Calibrating the Balance}
\label{sec:weight}

Once the decay is set, the weight $w$ determines how the energy trades off
errors in individual arcs ($\wSHD$) against errors in the interventional effects
($\wSID$). We first explain why the calibration rules for $\beta$ give no
guidance on $w$, and then use the first shell to derive the balance point at
which the two terms carry comparable weight.

Neither the shell rule \mref{eq:ballrule} nor the absolute rule of
Appendix~\ref{app:absolute} can guide the choice of $w$ for sparse DAGs, because
both are largely determined by arc additions, on which $\wSID$ vanishes and
$\wSHD$ depends only on the local decay $\beta w$. At $w = 0$, LGE assigns the
same prior probability to all $\G \supseteq \G_0$ and does not penalise spurious
arcs or overly dense networks. At $w = 1$, it reduces to the SHD-only priors of
Section~\ref{sec:lge}, which do not penalise differences in interventional
effects.

We can use the first shell $S_1(\G_0)$ to find the \emph{balance point} $w^* \in
(0, 1)$, the weight that gives comparable importance to $\wSHD$ and $\wSID$. For
every $\G \in S_1(\G_0)$, $w \wSHD(\G, \G_0) = w / |A_0|$ and $(1 - w) \wSID(\G,
\G_0) \leqslant 2(1 - w) / N$. Equating the first with the upper bound on the
second gives
\begin{equation}
  w^* = \frac{2 |A_0|}{N + 2 |A_0|} = \frac{2 \bar{d}}{1 + 2 \bar{d}},
  \qquad \bar{d} = \frac{|A_0|}{N},
\label{eq:wstar}
\end{equation}
where $\bar{d}$ is the average in-degree of $\G_0$ ($\bar{d} = 1$ and $w^* =
2/3$ in Figure~\ref{fig:example}). For \mbox{$w > w^*$}, $\wSHD$ dominates the
energy of every single arc edit, and LGE increasingly behaves like a purely
local prior. For $w < w^*$, $\wSID$ dominates and penalises omissions and
reversals more than spurious arcs. At $w = w^*$, a deletion costs at most twice
what an addition costs. Note that $(1 - w) \wSID(\G, \G_0) = 2(1 - w)/N$ only in
the worst case and \mbox{$\wSID(\G, \G_0) = 0$} on most of $S_1(\G_0)$: the
value at which $\wSHD$ and $\wSID$ balance exactly will typically be somewhat
smaller than $w^*$.

\subsection{Robustness to Errors in the Prior DAG}
\label{sec:robustness}

LGE relies on $\G_0$ to inform Bayesian estimation, but a prior DAG assembled
from an annotation database may contain arcs that are spurious, missing or
misoriented. Let the \emph{true DAG} underlying the data generating process be
$\G_T \neq \G_0$, and denote the \emph{perturbation level} of $\G_0$, the number
of node pairs at which it is wrong, as \mbox{$m = \SHD(\G_T, \G_0)$}. Then, the
robustness of the LGE prior is determined by how much prior mass is guaranteed
to remain on $\G_T$ as a function of $\beta$ and $w$. We first bound SID as a
function of the perturbation level (Lemma~\ref{theo:sidperturb}), then use that
bound to find the decay that maximises the prior probability guaranteed to
$\G_T$ (Theorem~\ref{theo:minimax}), and finally derive how much data are needed
to overturn a wrong prior DAG (Corollary~\ref{theo:evidence}).

Regardless of how $\G_T$ differs from $\G_0$, $\wSHD(\G_T, \G_0) = m / |A_0|$
and $\wSID(\G_T, \G_0)$ is bounded by the following lemma.

\begin{lemma}
  Let $\G, \G_0$ be DAGs over the same nodes, let $m = \SHD(\G, \G_0)$ and
  let $\PD(\G, \G_0)$ be the number of nodes at which they have different parent
  sets. Then
  \begin{equation*}
    \SID(\G, \G_0) \leqslant (N - 1) \PD(\G, \G_0),
    \qquad
    \PD(\G, \G_0) \leqslant \min\left\{N, 2m\right\},
  \end{equation*}
  and therefore $\wSID(\G, \G_0) \leqslant \sigma_m = \min\{1, 2m/N\}$.
\label{theo:sidperturb}
\end{lemma}
\begin{proof}
  See Appendix~\ref{app:robust}.
\end{proof}

At $m = 1$, Lemma~\ref{theo:sidperturb} gives $\SID(\G, \G_0) \leqslant 2(N -
1)$, the bound of \citet{peters}. If $\G_0 \subseteq \G_T$, so that the prior
DAG errs only by omission, then $\wSID(\G_T, \G_0) = 0$, a fact we will use in
Section~\ref{sec:sims}. The following theorem turns this bound into a lower
bound on the prior probability of $\G_T$.

\begin{theorem}
  Let $q = \exp\left\{-\beta w / |A_0|\right\}$ and
  $\sigma_m = \min\{1, 2m/N\}$ as in Lemma~\ref{theo:sidperturb}. Write
  $m_w = m + (1 - w)\sigma_m|A_0|/w$.
  Then, for every DAG $\G_T$ with $\SHD(\G_T, \G_0) = m$,
  \begin{equation}
    \Prob(\G_T) \geqslant \exp\left\{ -\beta\left[
      \frac{wm}{|A_0|} + (1 - w)\sigma_m \right] - N(N - 1) q \right\},
  \label{eq:robust}
  \end{equation}
  and, provided $m_w \leqslant N(N - 1)$, the right-hand side is maximised at
  \begin{equation}
    \beta^*(m) = \frac{|A_0|}{w} \log \frac{N(N - 1)}{m_w},
    \,\text{giving}\,
    \Prob(\G_T) \geqslant
      \exp\left[ -m_w\left\{ 1 + \log \frac{N(N - 1)}{m_w} \right\} \right].
  \label{eq:betastar}
  \end{equation}
\label{theo:minimax}
\end{theorem}
\begin{proof}
  See Appendix~\ref{app:robust}.
\end{proof}

In Proposition~\ref{theo:minimax-upper} (Appendix~\ref{app:robust}), we bound
from above the prior probability that any $\beta$ can guarantee to every such
$\G_T$. This upper bound is close to \mref{eq:betastar} at $w = 1$, but the gap
widens quickly as $w$ decreases, because \mref{eq:betastar} uses the worst case
of Lemma~\ref{theo:sidperturb}: for $w < 1$, the guarantee is conservative.
Three considerations follow.
\begin{itemize}
  \item We call $\beta^*(m)$ in \mref{eq:betastar} the \emph{robust decay}. It
    has the form of the shell rule \mref{eq:ballrule}, with $|S_1(\G_0)|$
    replaced by its upper bound $N(N - 1)$ and the mass ratio $r$ replaced by
    $m_w$, the perturbation level inflated by $\wSID$. It decreases in $m$, so a
    prior DAG believed to contain more errors calls for a slower decay, and
    pushing $\beta$ past $\beta^*(m)$ reduces the guarantee exponentially in
    $m$. For instance, if the true DAG is $\G_3$ in Figure~\ref{fig:example},
    then $m = 2$, $\sigma_2 = 2/3$ and, at $w = 0.5$, $m_w = 6$ and $\beta^*(2)
    = 12 \log 5 \approx 19.3$, half the decay that the shell rule suggests for a
    correct $\G_0$.
 \item Putting weight on $\wSID$ makes the prior less forgiving of errors in
   $\G_0$. When $2m \leqslant N$, $m_w = m\kappa_w$ with the \emph{inflation
   factor} $\kappa_w = 1 + 2\bar{d}(1 - w)/w$ and $\bar{d} = |A_0|/N$: in
   the worst case, a prior DAG with $m$ wrong arcs is penalised as if it had
   $m\kappa_w$ of them, because each wrong arc also creates interventional
   errors. There is no inflation at $w = 1$, the error count
   doubles at the balance point $w^*$ of \mref{eq:wstar}, and it grows without
   bound as $w \to 0$, so robustness to a wrong $\G_0$ favours a larger $w$. The
   exception is a $\G_0$ whose errors are all omissions: then $\wSID(\G_T,
   \G_0) = 0$ and $w$ is irrelevant.
  \item $Z(\beta, w)$ cancels from the prior odds of $\G_T$ against $\G_0$,
    and $E(\G_0, \G_0; w) = 0$, so
    \begin{equation*}
      \log \frac{\Prob(\G_T \given \D)}{\Prob(\G_0 \given \D)} =
      \log \frac{\Prob(\D \given \G_T)}{\Prob(\D \given \G_0)} -
      \beta E(\G_T, \G_0; w).
    \end{equation*}
    The prior adds a fixed penalty $\beta E(\G_T, \G_0; w)$ that does not
    depend on $n$: the posterior prefers $\G_T$ only once the Bayes factor in
    its favour exceeds $\exp\{\beta E(\G_T, \G_0; w)\}$, rather than $1$. The
    corollary below combines this with Lemma~\ref{theo:sidperturb} to find how
    much data that requires. Its assumption on the Bayes factor holds whenever
    $\log \Prob(\D \given \G)$ equals the maximised log-likelihood up to terms
    of smaller order than $n$, which is the case for the marginal likelihood
    and for the scores that approximate it, such as the Bayesian Information
    Criterion \citep[BIC;][]{bic}.
\end{itemize}

\begin{corollary}
  Let $\sigma_m$ and $m_w$ be as in Theorem~\ref{theo:minimax}, and let $\D$ be
  a sample of $n$ observations from a probability distribution $p_T$ that
  factorises according to $\G_T$ as in \mref{eq:parents}. Assume that
  \begin{equation*}
    \frac{1}{n} \log \frac{\Prob(\D \given \G_T)}{\Prob(\D \given \G_0)}
      \longrightarrow \Delta \geqslant 0 \qquad \text{in probability,}
  \end{equation*}
  where $\Delta$ is the smallest Kullback-Leibler divergence between $p_T$ and
  the distributions that factorise according to $\G_0$; $\Delta > 0$ if and
  only if $p_T$ cannot be written in the factorised form of $\G_0$. Then:
  \begin{enumerate}
    \item If $\Delta > 0$, $\Prob(\G_T \given \D) / \Prob(\G_0 \given \D)
      \to \infty$ in probability as $n \to \infty$. Approximating to
      the first order, the posterior switches from $\G_0$ to $\G_T$ when $n$
      exceeds
      \begin{equation*}
        \frac{\beta E(\G_T, \G_0; w)}{\Delta} \leqslant
          \frac{\beta}{\Delta} \left[\frac{wm}{|A_0|} + (1 - w)\sigma_m\right],
      \end{equation*}
      which at the robust decay $\beta^*(m)$ is at most
      $m_w \log\{N(N - 1)/m_w\} / \Delta$.
    \item If $\G_T$ and $\G_0$ are Markov equivalent,
      $\Prob(\D \given \G_T) = \Prob(\D \given \G_0)$ as noted in
      Section~\ref{sec:notation}, so the Bayes factor equals $1$ for every $n$
      and the posterior prefers $\G_0$ whenever \mbox{$E(\G_T, \G_0; w) > 0$},
      which holds for every $w > 0$ and $\G_T \neq \G_0$.
  \end{enumerate}
\label{theo:evidence}
\end{corollary}
\begin{proof}
  See Appendix~\ref{app:robust}.
\end{proof}

When $\Delta = 0$ because $\G_0$ contains every arc of $\G_T$ plus some
spurious ones, the Bayes factor grows only logarithmically in $n$
(Appendix~\ref{app:robust}). Hence, unless $\G_T$ and $\G_0$ are Markov
equivalent, enough data dominate the prior at any fixed decay, while at a fixed
sample size the prior dominates the data if the decay is fast enough.

Appendix~\ref{app:numerical} evaluates $\beta^*(m)$ on the reference networks
at the perturbation levels of Section~\ref{sec:sims}; and
Appendix~\ref{app:robust} checks Lemma~\ref{theo:sidperturb},
Theorem~\ref{theo:minimax} and Proposition~\ref{theo:minimax-upper} against the
enumeration of Appendix~\ref{app:enumeration}.

\section{Experiments}
\label{sec:experiments}

We now evaluate LGE empirically. After summarising how the calibration rules of
Section~\ref{sec:lge} behave on the four reference Gaussian networks from
Table~\ref{tab:gaussian}, we study the structural accuracy of the MAP DAGs
learned with LGE from simulated data with correct and perturbed prior DAGs
(Section~\ref{sec:sims}). Finally, we apply it to the reconstruction of a gene
regulatory network from single-cell kidney data (Section~\ref{sec:kidney}). All
experiments use the \textbf{bnlearn} R package \citep{jss09}.

Firstly, we computed the calibration rules of Section~\ref{sec:lge} on the four
reference networks, taking their DAGs as the prior $\G_0$; the details are in
Appendix~\ref{app:numerical} of the Supplementary Material. Three results guide
the choice of $\beta$ and $w$ in the rest of this section.
\begin{itemize}
  \item \emph{The balance point is nearly the same for all four networks.} The
    weight $w^*$ from \mref{eq:wstar}, which balances $\wSHD$ and $\wSID$,
    depends on $\G_0$ only through its average in-degree $\bar{d}$. The four
    DAGs are similarly sparse ($\bar{d} \approx 1.5$), so $w^*$ only ranges from
    $0.737$ to $0.761$ (Table~\ref{tab:gaussian}). Hence, the same weights can
    be used for all the networks.
  \item \emph{The decay prescribed by the shell rule is reliable, but
    network-specific.} The local decay $\beta w$ given by the shell rule
    \mref{eq:ballrule} with $r = 1$ scales with $|A_0|$, so it differs
    markedly between the networks: from $495$ for MAGIC-NIAB to $1396$ for
    ARTH150 (Table~\ref{tab:calibration} in Appendix~\ref{app:numerical}). Two
    checks confirm the accuracy of \mref{eq:ballrule}. The rule bounds the mass
    of $S_1(\G_0)$ from above; the exact mass, computed from the SID of each
    DAG, falls within $6\%$ of the bound. Furthermore, controlling all the
    shells, as in Appendix~\ref{app:absolute}, changes $\beta$ by at most $5\%$
    across the four networks, confirming that the first shell is sufficient to
    achieve an accurate calibration.

  \item \emph{A prior DAG that may be wrong calls for a slower decay.} The
    robust decay $\beta^*(m)$ from \mref{eq:betastar} maximises the prior mass
    guaranteed to the true DAG when $\G_0$ is wrong at $m$ node pairs, and it
    decreases in $m$. On ECOLI70 with $w = 0.5$, it is $619$, $527$ and $429$
    when $10\%$, $20\%$ and $50\%$ of the arcs of $\G_0$ are perturbed,
    against $\beta = 1058$ from the shell rule for a correct $\G_0$. Hence,
    the more errors we suspect in $\G_0$, the slower the decay should be, at
    the price of a prior that is less concentrated around $\G_0$ when it is in
    fact correct.
\end{itemize}

\begin{table}[b]
\begin{center}
  \begin{tabular}{lrrrrr}
  \cline{1-6}
  \rule{0pt}{2.6ex}network & $N$ & $|A_0|$ & $|\Theta|$ & $\bar{d}$ & $w^*$ \\
  \cline{1-6}
  ARTH150    & $107$ & $150$ & $364$ & $1.40$ & $0.737$ \\
  ECOLI70    & $46$  & $70$  & $162$ & $1.52$ & $0.753$ \\
  MAGIC-IRRI & $64$  & $102$ & $230$ & $1.59$ & $0.761$ \\
  MAGIC-NIAB & $44$  & $66$  & $154$ & $1.50$ & $0.750$ \\
  \cline{1-6}
  \end{tabular}
  \caption{The Gaussian networks from the BN repository \citep{bnrepo}, with
    their numbers of nodes ($N$), arcs ($|A_0|$), parameters ($|\Theta|$);
    average in-degrees $\bar{d} = |A_0| / N$, the $w^*$ from \mref{eq:wstar}.}
  \label{tab:gaussian}
\end{center}
\end{table}

\subsection{Simulations}
\label{sec:sims}

Our simulation study answers six questions.
\begin{enumerate}
  \item[Q1.] How does the accuracy of LGE change with the decay $\beta$ when
    $\G_0$ is correct, and how does LGE compare with the uniform prior?
  \item[Q2.] Does LGE also outperform the marginal uniform prior?
  \item[Q3.] How sensitive is LGE to the choice of the weight $w$?
  \item[Q4.] Does including SID in the energy improve on using SHD alone, as
    the existing informative priors do?
  \item[Q5.] How robust is LGE to errors in $\G_0$?
  \item[Q6.] Do the improvements extend to measures of accuracy that LGE does
    not optimise?
\end{enumerate}
The study spans the four Gaussian reference networks of
Table~\ref{tab:gaussian}, which were learned from gene expression (ARTH150,
ECOLI70) or sequence (MAGIC-NIAB, MAGIC-IRRI) data and are therefore
representative of the domains that motivated this paper. Furthermore, the
marginal likelihood of Gaussian BNs, BGe \citep{bge}, has a closed form: it is
computed exactly and its properties are well known. This is not the case for
more complex distributional assumptions, such as the zero-inflated models of
\citet{zigdag} we use in Section~\ref{sec:kidney}, where the marginal likelihood
must be replaced by an approximation such as BIC. Hence, we can
evaluate the LGE prior without the results being affected by errors in the
estimation of the marginal likelihood. From each network, we generated $10$ data
sets for each of the sample sizes $n/|\Theta| = 1, 2, 5, 10, 20, 50$ to align
results across BNs of different complexity.

From each data set, we learned a MAP DAG with tabu search \citep{glover} and the
BGe marginal likelihood with its default hyperparameters \citep{bge}. We used
the following priors:
\begin{itemize}
  \item The (uninformative) uniform prior, which is the LGE prior with
    $\beta = 0$.
  \item The (weakly informative) marginal uniform prior from \citet{pgm16},
    with an arc inclusion probability giving an expected density of 3 arcs per
    node.
  \item The LGE prior, over the grid $\beta = \{2, 50, 200, 500\}$ and
    $w = \{0, 0.25, 0.50, 0.75, 1\}$, together with $\beta = 1000$ at $w = 1$.
\end{itemize}
For $\G_0$, we used the DAG of the reference BN, either unchanged
or with a number of perturbed arcs equal to $0.1|A_0|$, $0.2|A_0|$ or
$0.5|A_0|$, each perturbation sampling an addition, a deletion or a reversal
with equal probability. The study comprises six arms:
\begin{itemize}
  \item The \emph{full grid} with an unperturbed $\G_0$, which shows how
    accuracy changes with $\beta$ and $w$ when $\G_0$ is correct (Q1 to Q3).
  \item A \emph{decay sweep} over $\beta$ at $w = 0.5$, with a perturbed $\G_0$
    (Q5).
  \item A \emph{weight sweep} over $w$ at $\beta = 50$, with a perturbed $\G_0$
    (Q4).
  \item A \emph{matched local decay} at each perturbation level: $(\beta, w) =
    (200, 0.25)$ against $(50, 1)$ from the weight sweep, which put the same
    penalty $\beta w = 50$ on spurious arcs but different weights on $\wSID$
    (Q4).
  \item The \emph{fastest decay} $(\beta, w) = (1000, 1)$, past the calibration
    threshold, with an unperturbed $\G_0$ and with $0.5|A_0|$ perturbations (Q1,
    Q5).
  \item A \emph{deletion-only} rerun of the matched pair: the omission-only case
    of Lemma~\ref{theo:sidperturb} (Q4).
\end{itemize}
The grid ranges over the values of $\beta$ and $w$ given by the calibration
rules of Sections~\ref{sec:calibration} to~\ref{sec:robustness} and checked in
Appendix~\ref{app:numerical}. Its lower half is a slow-decay regime in which
the prior is not concentrated around $\G_0$; $\beta w = 1000$ clears the
threshold of the shell rule \mref{eq:ballrule} with $r = 1$
(Table~\ref{tab:calibration}) on three of the four networks, and $\beta = 500$
falls within the range of the robust decay $\beta^*(m)$ of \mref{eq:betastar}
across the networks and perturbation levels.

\begin{figure}[p]
  \includegraphics[width=\linewidth]{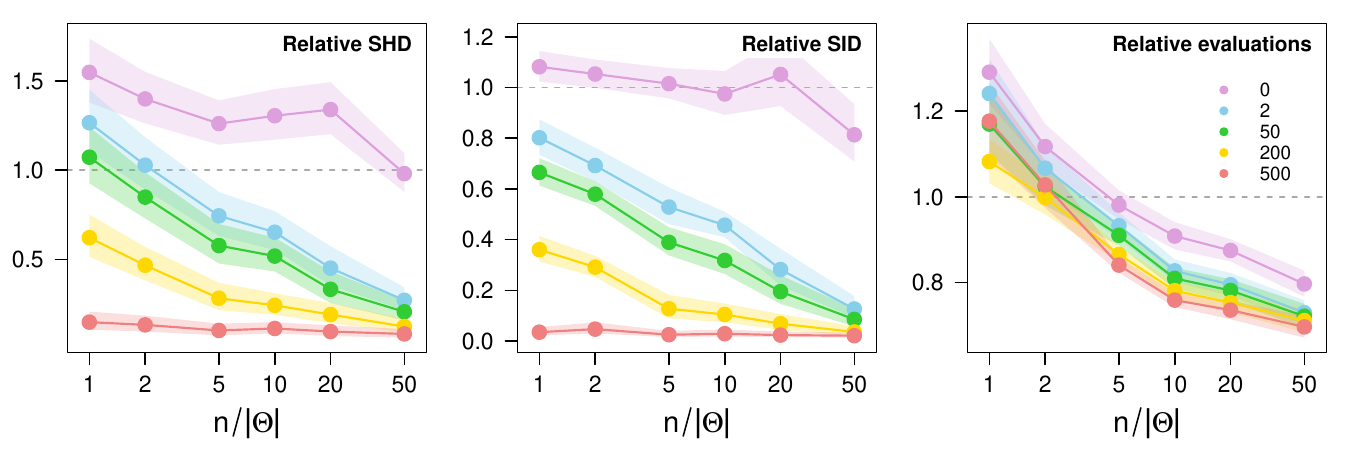}
  \caption{Performance for various $\beta$ relative to the marginal uniform
    prior, with an unperturbed $\G_0$. Points are geometric means of the paired
    per-run ratios; bands are $90\%$ confidence intervals on the log scale.}
  \label{fig:ep-decay}

  \vspace{1.5em}
  \includegraphics[width=\linewidth]{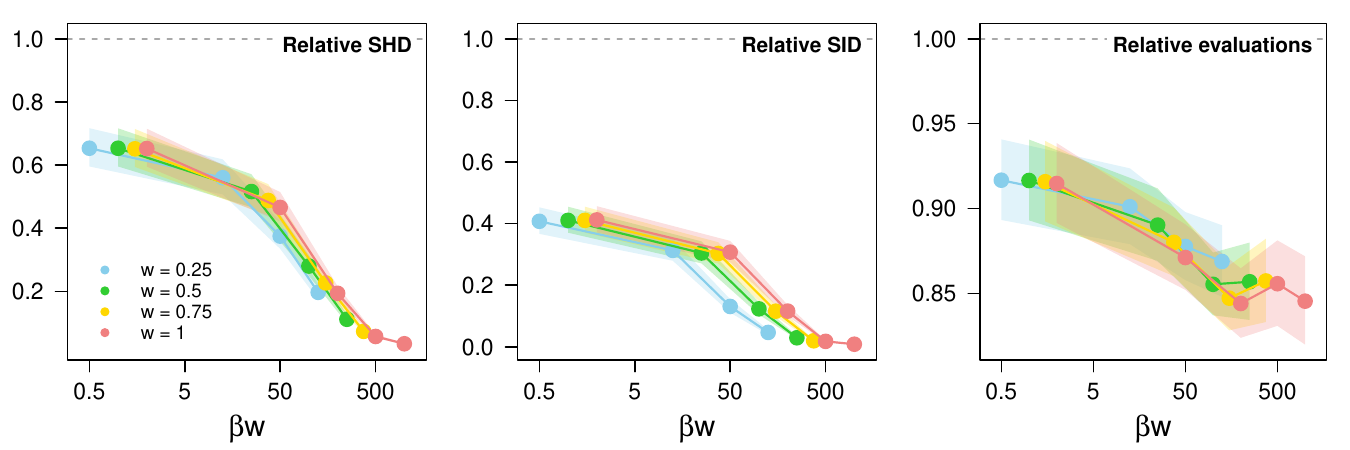}
  \caption{Performance for various $w$ against the local decay $\beta w$,
    relative to the marginal uniform prior, with an unperturbed $\G_0$ (90\%
    confidence bands). The runs with $w = 0$ are omitted: $\wSHD$ drops out of
    the energy, leaving no local decay to match on.}
  \label{fig:ep-weights}

  \vspace{1.5em}
  \includegraphics[width=\linewidth]{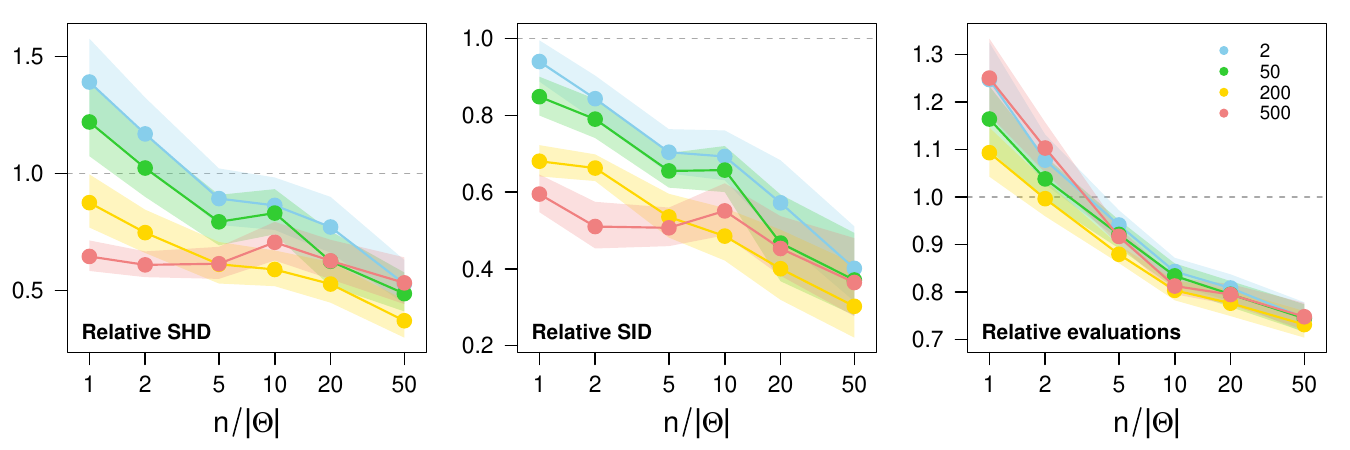}
  \caption{As Figure~\ref{fig:ep-decay}, with $0.5|A_0|$ perturbations (90\%
    confidence bands).}
  \label{fig:ep-perturb}
\end{figure}

\begin{figure}[t]
  \includegraphics[width=\linewidth]{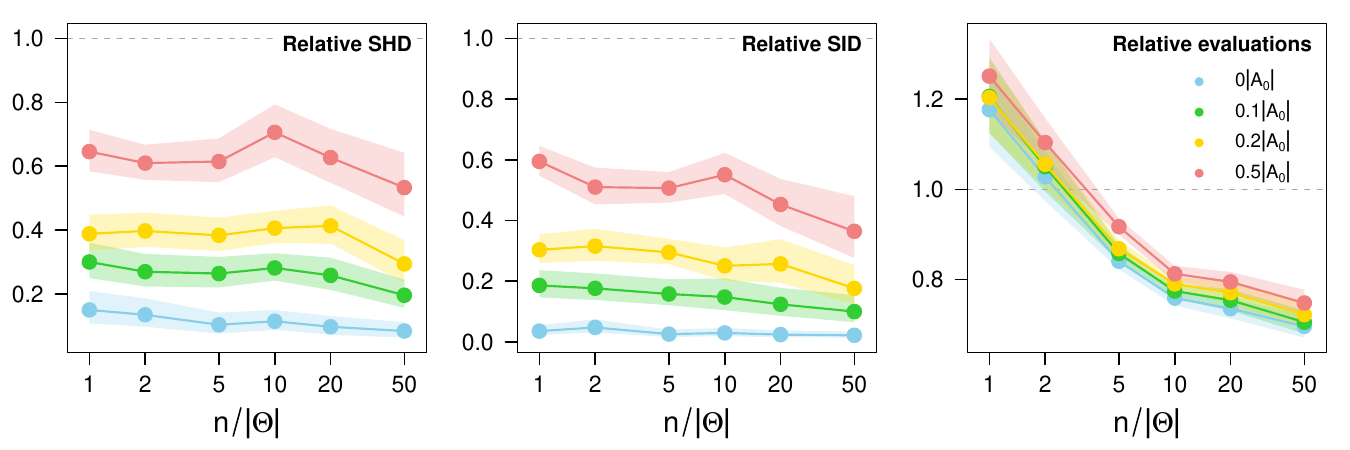}
  \caption{Performance for various perturbation levels of $\G_0$ relative to
    the marginal uniform prior, at $\beta = 500$ and $w = 0.5$ (90\% confidence
    bands).}
  \label{fig:ep-npert}
\end{figure}

We measure structural accuracy with $\wSHD$ and $\wSID$, and speed with the
number of evaluated local distributions. We present each of them relative to a
baseline prior through paired comparisons: for each data set $d$, we divide the
value $M_{\mathrm{LGE}}(d)$ of the metric for the DAG learned with LGE by the
value $M_{\mathrm{base}}(d)$ for the DAG learned from the same data set with
the baseline prior. Pairing removes the confounding effects of the reference
networks and of the data sets: some networks are harder to learn than others,
and some data sets are less informative than others, but these differences
affect the evaluation of both priors in the same way. Furthermore, $\wSHD$,
$\wSID$ and $n/|\Theta|$ are all rescaled for model complexity, so the ratios
can be aggregated across networks. We summarise the ratios over the $D$ data
sets that correspond to each point in Figures~\ref{fig:ep-decay}
to~\ref{fig:ep-npert} with their geometric mean
\begin{equation*}
  \exp\left\{ \frac{1}{D} \sum_{d = 1}^{D}
    \log \frac{M_{\mathrm{LGE}}(d) + \delta}
              {M_{\mathrm{base}}(d) + \delta} \right\},
\end{equation*}
where $\delta$ is a small constant that keeps the ratios defined when a
divergence is zero. The ratios are heavily skewed, because the baseline can be
close to zero, and their arithmetic mean would be dominated by a few extreme
values; the geometric mean is more robust. The bands in the figures are $90\%$
confidence intervals computed on the same log scale. In addition, we measure the
Kullback-Leibler divergence $\KL(\G, \G_T)$ \citep{kullback} from the true DAG
and the optimal adjustment distance of \citet{henckel}, two measures of accuracy
that LGE is not a function of (Q6).

\emph{Q1 (decay and the uniform prior).} Figure~\ref{fig:ep-decay} shows LGE
with $w = 0.5$ and an unperturbed $\G_0$ for each $\beta$ in the grid, including
$\beta = 0$, the uniform prior, all relative to the marginal uniform prior:
values below $1$ mark priors that outperform it. The uniform prior is the worst
of the three: every $\beta > 0$ improves on it. LGE's accuracy improves
monotonically in $\beta$: the more concentrated the prior is around a correct
$\G_0$, the closer the learned DAGs are to $\G_T$ in both $\wSHD$ and $\wSID$.
The improvement plateaus once the calibration rules are met. In the fastest
decay arm ($\beta w = 1000$, $w = 1$), LGE recovers $\G_T$ exactly in $87\%$ to
$100\%$ of the runs on the three networks for which the shell rule with $r = 1$
gives $\beta w$ between $495$ and $841$ (Table~\ref{tab:calibration}). It does
so in only $13\%$ of the runs on ARTH150, for which the shell rule gives $\beta
w = 1396$.

\emph{Q2 (the marginal uniform prior).} The marginal uniform prior assigns the
same probability to each of the two possible arcs between each node pair
(Section~\ref{sec:notation}). It is weakly informative, since it only controls
the sparsity of the learned DAGs, but \citet{pgm16} showed in simulations that
it improves their structural accuracy over the uniform prior. LGE does
outperform it: in Figure~\ref{fig:ep-decay}, LGE improves on the marginal
uniform prior everywhere except at the slowest decays ($\beta = 2, 50$) and the
smallest sample sizes. There, LGE is close to the uniform prior and, with little
data to correct it, shares its tendency to favour dense DAGs. On ARTH150,
improvements are visible from $\beta = 200$ onward, and clearly only at $\beta =
500$, in line with its larger calibration threshold (Q1). LGE also evaluates
fewer local distributions than the marginal uniform prior as $\beta$ grows:
about $30\%$ fewer at $n/|\Theta| = 50$, although $18\%$ more at $n/|\Theta| =
1$, where convergence to the MAP DAG is slower. Note that no prior over DAGs can
outperform another uniformly: which performs better depends on the DAG
underlying the data. Since $\G_0 = \G_T$ in Q1 and Q2, they show how much LGE
can gain from a correct $\G_0$; Q5 investigates how much of that gain survives
errors in $\G_0$.

\emph{Q3 (the weight $w$).} Comparing different weights at the same $\beta$
would be misleading: the shell rule \mref{eq:ballrule} shows that the strength
of the prior depends on the local decay $\beta w$, so a larger $w$ also makes
the prior stronger. Therefore, Figure~\ref{fig:ep-weights} draws one curve for
each weight against $\beta w$, with the points on each curve corresponding to
the $\beta$ of the grid in increasing order. At the same $\beta w$, all the
priors penalise each spurious arc in the same way. The curves coincide on SHD
and on the number of evaluations, and differ only for SID, where $w = 0.25$ is
ahead from $\beta w = 50$ onward. Hence, at a matched local decay, the choice of
$w$ only affects how well LGE recovers the interventional effects encoded in
$\G_0$: putting more weight on $\wSID$ improves SID without making SHD or the
speed of the search any worse.

\emph{Q4 (SID against the SHD-only priors).} Section~\ref{sec:lge} showed that
LGE with $w = 1$ coincides with the SHD-only priors of \citet{imoto04} and
\citet{csprior} built from $\G_0$; at $\beta = 50$, the latter puts probability
$\pi$ between $0.41$ (ARTH150) and $0.52$ (MAGIC-NIAB) on each state of
$\G_0$. Hence, we are effectively comparing LGE against the existing informative
priors rather than a weakly informative one. The matched local decay arm of our
simulation study provides the two points $(\beta, w) = (200, 0.25)$ and $(50,
1)$ that correspond to $\beta w = 50$ in Figure~\ref{fig:ep-weights}. At that
matched local decay, the SID-heavy prior is $20\%$ more accurate in terms of
$\wSHD$ and $57\%$ in terms of $\wSID$, dominating the $w = 1$ case in all $24$
combinations of network and sample size and in $80\%$ ($\wSHD$) and $85\%$
($\wSID$) of the individual runs. Hence, for the same penalty on each spurious
arc, also penalising the interventional effects that $\G$ fails to identify
produces DAGs that are closer to $\G_T$ in both their arcs and their causal
effects. The same holds with a perturbed $\G_0$: at $\beta w = 50$, the
SID-heavy prior is ahead in $23$, $24$ and $22$ of the $24$ combinations of
network and sample size at $10\%$, $20\%$ and $50\%$ perturbation, by a margin
that shrinks as the perturbation grows. When the perturbations are all deletions
(the deletion-only arm), the omission-only case of Lemma~\ref{theo:sidperturb}
in which $\wSID(\G_T, \G_0) = 0$, the margin vanishes at $0.5|A_0|$.

\emph{Q5 (errors in $\G_0$).} In practice, $\G_0$ will contain spurious,
missing or misoriented arcs. To study their effect, we perturbed $\G_0$ by
adding, deleting or reversing $10\%$, $20\%$ or $50\%$ of its arcs, so that
$\G_0 \neq \G_T$. We compare the learned DAGs both with the marginal uniform
prior, as above, and with the perturbed $\G_0$ itself, which tells us whether
the data correct the errors in $\G_0$ or the prior carries them over.
Figure~\ref{fig:ep-perturb} repeats Figure~\ref{fig:ep-decay} with $0.5|A_0|$
perturbations. LGE still improves on the marginal uniform prior at every decay,
except for $\beta = 2, 50$ at $n/|\Theta| = 1, 2$. Compared with the perturbed
$\G_0$ itself, the learned DAGs are $25\%$ closer to $\G_T$ in $\wSHD$ and
$44\%$ closer in $\wSID$ at $\beta = 200$ and $500$, from $n/|\Theta| = 5$
onward. However, larger values of $\beta$ no longer yield the largest
improvement: with a correct $\G_0$, the curves are ordered by $\beta$ at every
sample size and never meet, while with $0.5|A_0|$ perturbations $\beta = 200$
starts behind $\beta = 500$, catches up with it at $n/|\Theta| = 5$ and passes
it from $n/|\Theta| = 10$. The fewer the errors in $\G_0$, the later this
happens: $\beta = 200$ catches up with $\beta = 500$ at $n/|\Theta| = 10$ with
$0.2|A_0|$ perturbations and only at $n/|\Theta| = 50$ with $0.1|A_0|$. This
confirms Corollary~\ref{theo:evidence}: the evidence needed to overturn an
incorrect $\G_0$ does not grow with $n$, so enough data dominate the prior at
any fixed decay, while at a fixed sample size the prior dominates the data if
the decay is fast enough. Calibrating $\beta$ to the errors expected in $\G_0$,
as suggested by \mref{eq:betastar}, matters most at smaller sample sizes. The
cost of ignoring this is visible in the fastest decay arm ($\beta w = 1000$):
with half of the arcs of $\G_0$ perturbed, LGE's $\wSHD$ on MAGIC-NIAB is $36\%$
to $53\%$ larger than that of the marginal uniform prior for $n/|\Theta|
\geqslant 10$, while on the other three networks it remains $20\%$ to $45\%$
smaller.

Figure~\ref{fig:ep-npert} shows how performance degrades as $\G_0$ becomes
increasingly incorrect, at $\beta = 500$ and $w = 0.5$, with one curve for each
perturbation level. In $\wSHD$ and $\wSID$, the curves are ordered by the
perturbation level at every sample size and never intersect, and the degradation
is gradual: relative to the marginal uniform prior, $\wSHD$ is $0.11$, $0.26$,
$0.38$ and $0.62$ and $\wSID$ is $0.03$, $0.15$, $0.26$ and $0.49$ at $0$,
$0.1|A_0|$, $0.2|A_0|$ and $0.5|A_0|$ perturbations. Even at $0.5|A_0|$, LGE
beats the marginal uniform prior in $\wSID$ in all $24$ combinations of network
and sample size, and in $\wSHD$ in $21$. Compared with the perturbed $\G_0$
itself, the learned DAGs are closer to $\G_T$ in $\wSID$ at every level, by
$31\%$, $38\%$ and $44\%$. In $\wSHD$, they are about a quarter closer at
$0.5|A_0|$, level at $0.2|A_0|$ and $26\%$ farther at $0.1|A_0|$: a $\G_0$ that
is $90\%$ right has fewer wrong arcs than the DAG the search returns with a
prior of this strength. LGE thus remains beneficial when $\G_0$ is a reasonable
guess of $\G_T$, and starts failing only with a poor $\G_0$ and a large $\beta$.

\emph{Q6 (measures LGE does not optimise).} $\KL(\G, \G_T)$ compares the
probability distributions of the learned and the reference BNs rather than their
DAGs, and the optimal adjustment distance identifies causal effects with
different adjustment sets than SID. Neither is a function of SHD or SID, so
improvements in them cannot be an artefact of the prior. We report them in
Appendix~\ref{app:kl} (Tables~\ref{tab:kl-decay} and~\ref{tab:kl-network}, and
Tables~\ref{tab:aid-mu} and~\ref{tab:aid-network}); the two highlight the same
trends. Against the uniform prior, LGE is ahead at every $\beta$, on every
network and at every sample size, with and without perturbation, and its
advantage grows with $\beta$. Against the marginal uniform prior, LGE falls
behind at $n/|\Theta| \leqslant 2$ for $\beta \leqslant 50$ (in $\KL(\G, \G_T)$,
and in the optimal adjustment distance with a perturbed $\G_0$), but outperforms
it at every $\beta$ from $n/|\Theta| = 5$ onward. At $\beta = 500$, LGE is the
best performer in $90\%$ ($\KL(\G, \G_T)$) and $100\%$ (optimal adjustment
distance) of the runs with a correct $\G_0$, and in $81\%$ and $87\%$ with a
perturbed one.

\emph{Overall}, LGE compares favourably with all the other priors. It
outperforms the uniform prior at every decay, weight and sample size, whether
$\G_0$ is correct or not. It outperforms the marginal uniform prior except at
the slowest decays with the smallest samples and, on one network, at a decay
much faster than the robust decay with half of the arcs of $\G_0$ wrong. With
$w < 1$, it outperforms the SHD-only informative priors from \citet{csprior}
and \citet{imoto04} at the same local decay.

\subsection{Kidney Gene Expression Network Reconstruction}
\label{sec:kidney}

We now apply LGE in the setting that motivated it: reconstructing a gene
regulatory network with a prior DAG built from an annotation database.
Specifically, we use the single-cell transcriptomic data of the human kidney
atlas of \citet{lake23}.\footnote{Available from the Gene Expression Omnibus
under accession number GSE183276.} We describe the data and the prior DAG,
explain how we calibrated LGE knowing that the prior DAG is incomplete, and
compare the DAGs learned with and without it. We restricted the analysis to the
$3087$ proximal tubule cells sampled from healthy reference tissue and annotated
as being in the reference state: a homogeneous population whose gene expression
is controlled by a few well-characterised transcription factors, such as HNF4A,
HNF1A and PPARA, which makes it a natural setting for an informative prior. From
these cells, we selected $50$ genes: $10$ transcription factors and $40$ genes
that the Human Protein Atlas \citep{hpa} reports as enriched in proximal tubular
cells. Single-cell data are sparse counts: $55\%$ are zero, and the
transcription factors are detected in only $7\%$ to $40\%$ of the cells.
Therefore, we modelled each gene with a zero-inflated negative binomial
distribution and used the corresponding BIC as the network score to replace the
marginal likelihood of the data, which is not available in closed form,
similarly to \citet{zigdag}. We learned MAP DAGs by greedy hill climbing. A
single MAP DAG learned from such sparse and noisy counts can change considerably
from one sample of cells to another. For this reason, we used model averaging to
ensure the robustness of our models \citep{aime11}: we learned a MAP DAG from
each of $50$ random subsamples of $2000$ cells, and kept the arcs that appear in
a large enough fraction of them, with the threshold estimated from the data.
Appendix~\ref{app:kidney} of the Supplementary Material gives all the details
needed to reproduce the analysis.

\begin{figure}[t]
  \includegraphics[width=0.8\linewidth]{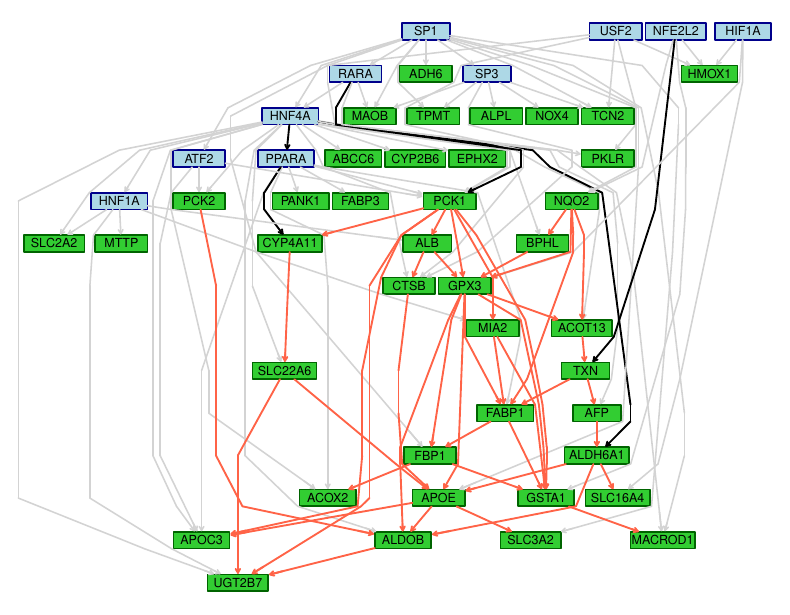}
  \caption{The posterior DAG learned from the proximal tubule single-cell data
    with LGE ($\beta = 670$, $w = 0.83$): transcription factors in blue,
    target genes in green. Arcs are black if the pair of genes is adjacent in
    both the prior DAG and the DAG learned without LGE, light grey if only in
    the former, and red if only in the latter.}
  \label{fig:sct}
\end{figure}

We built the prior DAG $\G_0$ using the interactions between transcription
factors and their targets reported in DoRothEA \citep{dorothea}, retrieved
through OmniPath \citep{omnipath}. DoRothEA grades each interaction with one of
five confidence levels, depending on the amount and type of evidence that
supports it. We kept the interactions at the two highest levels between pairs of
genes among the $50$ we selected (Appendix~\ref{app:kidney} lists the criteria).
These interactions contain cycles, so we removed a small set of arcs that breaks
all of them (a feedback arc set) to obtain a DAG. The resulting $\G_0$ has
$|A_0| = 69$ arcs, all of them leaving one of the $10$ transcription factors
(the blue boxes in Figure~\ref{fig:sct}). Sequencing depth, the number of
transcripts captured in each cell, inflates the counts of all genes at once and
may confound their relationships. We therefore added it (on a log scale) as an
auxiliary root node that is a parent of every gene, which plays the role of the
depth covariate in the regression models of \citet{sctransform} and prevents
spurious arcs between highly expressed genes. The node is omitted from
Figures~\ref{fig:sct} and~\ref{fig:sct-all} for clarity.

We calibrated LGE using the rules from Section~\ref{sec:lge}, allowing for
$\G_0$ being incomplete. The two highest confidence levels are a conservative
selection of the interactions in DoRothEA, and $\G_0$ contains no arcs between
the $40$ target genes, so the network of molecular pathways linking these $50$
genes is likely to have more arcs than those in $\G_0$. Furthermore, we removed
some known arcs via the feedback arc set, and DoRothEA only reports arcs going
from transcription factors to genes. Hence, we calibrated $\beta$ and $w$ for a
prior DAG with $|A_0| = 120$ arcs, for which $\bar{d} = 2.4$ and \mref{eq:wstar}
gives the balance point $w^* = 0.83$, the value we used for $w$. Since we do not
know how many of the arcs of $\G_0$ are incorrect, we considered four regimes,
matching ealier perturbation levels: $0|A_0|$, $0.1|A_0|$, $0.2|A_0|$,
$0.5|A_0|$ incorrect arcs. The corresponding decays are $\beta = 1122$ from the
shell rule \mref{eq:ballrule}, and the robust $\beta^*(m) = 670$, $570$, $487$
from \mref{eq:betastar}.

Figure~\ref{fig:sct} shows the DAG learned with $\beta = 670$ and $w = w^*$;
Figure~\ref{fig:sct-all} in Appendix~\ref{app:kidney} shows those learned with
all four values of $\beta$, together with $\G_0$ and the DAG learned without
using the LGE prior (equivalently, with the uniform prior). The overlap between
the latter and $\G_0$ is limited: only $5$ of the $69$ arcs of $\G_0$ appear in
it. Only NFE2L2 $\rarr$ TXN and HNF4A $\rarr$ PPARA have the same direction. The
other $3$ arcs are reversed, with a target gene as the parent of a transcription
factor: CYP4A11 $\rarr$ PPARA, ALDH6A1 $\rarr$ HNF4A and PCK1 $\rarr$ RARA. The
first reversal is biologically plausible: CYP4A11 and PPAR$\alpha$ regulate each
other, forming a feedback loop \citep{lasker,ng07}. The other two have no
comparable support in the literature. Most of the remaining $93$ arcs connect
target genes, forming hubs around PCK1 and ALDH6A1. Such hubs cannot appear in
$\G_0$, since DoRothEA only describes the regulation of genes by transcription
factors. The two DAGs also disagree on the order of the genes
(Appendix~\ref{app:kidney:orderings}).

The posterior DAGs move from $\G_0$ towards the data-driven DAG as $\beta$
decreases. All contain every arc of $\G_0$ with the same direction, except for
one arc at $\beta = 487$. None connects a pair of genes that is adjacent
neither in $\G_0$ nor in the data-driven DAG. The value of $\beta$ determines
how many arcs from the data-driven DAG are included: from $38$ at $\beta =
1122$ to $53$ at $\beta = 570$ and $52$ at $\beta = 487$. In practice, the
prior replaces between $45$ and $60$ arcs of the data-driven DAG with the $63$
or $64$ arcs of $\G_0$ that the data-driven DAG lacks, removing most of the
hubs among the target genes. The topological orderings of the posterior DAGs
follow the same pattern, moving away from that of $\G_0$ and towards that of
the data-driven DAG as $\beta$ decreases.

Overall, after accounting for sequencing depth, the data carry little
information about the regulatory arcs in DoRothEA. LGE adds them, drops about
half of the arcs learned from the data alone, mostly hubs among target genes,
and adds no arc that is in neither network. Greater agreement with $\G_0$ is not
independent evidence of accuracy, since $\G_0$ is itself prior knowledge; what
LGE provides is a principled way to combine known regulatory relationships that
cannot be recovered from these data with novel ones that are not documented in
DoRothEA.

\section{Discussion and Conclusions}
\label{sec:conclusions}

In this paper, we proposed what is, to the best of our knowledge, the first
causal graphical prior for DAG learning, which we call the \emph{local-global
energy prior} (LGE). Motivated by the increasing availability of annotation
databases in the life and clinical sciences, we constructed it as the maximum
entropy prior for a given expected divergence from a prior DAG built from those
databases, measuring that divergence by two properties of a DAG that matter in
use: its local structure (SHD) and its causal pathways (SID).
Nevertheless, LGE has a general construction and can be applied to
other fields as well.

LGE is a causal graphical prior in the sense that it encodes the direct and
indirect causal relationships of the prior DAG $\G_0$ and transfers them to the
posterior $\Prob(\G \given \D)$. A score-equivalent score, such as BGe, cannot
distinguish between the DAGs in an equivalence class; LGE disambiguates them
using $\G_0$ as a guide. With a score that identifies causal structure through
its distributional assumptions, such as the zero-inflated models of
\citet{zigdag} we used in Section~\ref{sec:kidney}, LGE augments the information
in the data, which dominates it at larger sample sizes; its effect is largest at
the sample sizes at which causal identifiability, which relies on asymptotic
guarantees, has not yet taken hold.

Overall, LGE improves both the structural accuracy of the MAP DAGs we learned
in our experiments and the fit of the corresponding BNs to the true
distribution when calibrated as follows:
\begin{enumerate}
  \item if we trust the prior DAG $\G_0$, choose the local decay $\beta w$ by
    the shell rule \mref{eq:ballrule} with mass ratio $r = 1$;
  \item if we do not, choose $\beta$ as the robust $\beta^*(m)$ from
    \mref{eq:betastar}, treating it as an upper bound;
  \item take $w$ at or just below the balance point $w^*$ of \mref{eq:wstar}.
\end{enumerate}
If $\G_0$ is known to be incomplete, as prior DAGs built from annotation
databases typically are, the rules should use the number of arcs expected in the
true DAG in place of $|A_0|$, as we did in Section~\ref{sec:kidney}. The rules
are sufficient rather than necessary, and LGE performed well whenever $\beta$
and $w$ were in the right order of magnitude, including when the prior DAG was
partially incorrect. In Section~\ref{sec:kidney}, it did so when combined with
the BIC of a zero-inflated negative binomial model, suggesting that LGE can be
combined with penalised-likelihood scores when the marginal likelihood is not
available in closed form.

Nevertheless, we note LGE's computational cost. On the networks of
Table~\ref{tab:gaussian}, computing SID in full on a candidate DAG takes
$0.05$ms, but that grows $O(N^3)$, which can become overwhelming for large
networks if SID is recomputed for every candidate. Caching, as used to scale
greedy search to large networks \citep{stco17}, contains that cost: the prior
DAG $\G_0$ is fixed, so its path matrix, the expensive part of computing SID
\citep{peters}, is computed once; and SID is a sum over nodes of terms that
depend on one parent set each, so a single-arc move changes at most two of them.
Caching those terms alongside the local scores reduces the cost of the prior to
that of a decomposable score. Replacing SID with a causal divergence designed
for efficient computation, such as those from \citet{henckel}, may improve speed
further, although it would also change the semantics of the prior.

We did not evaluate LGE in MCMC posterior sampling, which is an order of
magnitude more expensive than greedy search. Nevertheless, we expect LGE to
work as well in a sampler as in MAP search. Structure MCMC moves between DAGs
by adding, deleting or reversing a single arc \citep{giusi-uai}, the same moves
as the search in Section~\ref{sec:sims}. It accepts a move with a probability
that depends on the prior only through the ratio $\Prob(\G') / \Prob(\G)$.
The normalising constant $Z(\beta, w)$ cancels out of this ratio, so the prior
is as cheap to evaluate in a sampler as in a search. Furthermore, the
calibration rules of Section~\ref{sec:lge} and Corollary~\ref{theo:evidence}
describe prior and posterior probabilities, not the DAG a search returns. Hence,
they hold for the posterior a sampler draws from. What remains open is how LGE
affects the spread of the posterior and the mixing of the sampler. A partially
wrong $\G_0$ gives the posterior two modes, one near $\G_0$ and one near $\G_T$.
This is the regime in which the decays cross in Figure~\ref{fig:ep-perturb}, and
a sampler would have to travel between the two modes.

Two limitations of LGE stem from $\G_0$ being a DAG. Firstly, it cannot encode
feedback loops. With time-course data, these can be represented by the unrolled
DAGs used to model dynamic BNs \citep{murphy02}; with snapshot data such as
those in Section~\ref{sec:kidney}, they cannot, and we had to remove a feedback
arc set from the interactions in DoRothEA to obtain $\G_0$. Secondly, it
cannot encode an arc whose direction is unknown \emph{a priori}: $\G_0$ is
required to be a completely directed graph, and both SHD and SID penalise arcs
whose directions differ from those in $\G_0$. A term in the energy that compares
the skeletons of $\G$ and $\G_0$ for such arcs would lift this restriction while
keeping the Gibbs form of the prior.

\vfill

\begin{funding}
Marco Scutari is partially supported by grant K25-2170 from the International
Joint Usage/Research Center, the Institute of Medical Science, the University of
Tokyo.
Shuichi Kawano is partially supported by JSPS KAKENHI Grant Numbers JP23H00809
and JP23H03352.
\end{funding}

\clearpage


\input{paper.bbl}
\clearpage
\pagebreak
\include{supplement}

\end{document}

%% file: supplement.tex

\begin{supplement}

\stitle{Supplementary Material}
\sdescription{The rescaling of SHD and SID; the normalising constant of the
  LGE prior, the calibration of its decay and its numerical evaluation on
  reference networks; the proofs of the robustness results from
  Section~\ref{sec:lge}; the Kullback-Leibler and adjustment-distance results
  from Section~\ref{sec:sims}; and the experimental setting of the single-cell
  analysis in Section~\ref{sec:kidney}.}

\appendix
\renewcommand{\thetable}{\Alph{section}\arabic{table}}
\renewcommand{\thefigure}{\Alph{section}\arabic{figure}}
\makeatletter
\@addtoreset{table}{section}
\@addtoreset{figure}{section}
\makeatother

\section{Rescaling SHD and SID}
\label{app:rescaling}

\subsection{Choice of the Rescaling}
\label{app:rescaling:choice}

The energy \mref{eq:energy} is a convex combination of SHD and SID, which take
values over very different ranges, so Section~\ref{sec:lge} rescales both to
make their convex combination meaningful. Here, we start from their natural
upper bounds, and then show why we divide SHD by the number of arcs $|A_0|$ of
the prior DAG and SID by $N(N - 1)$. Throughout, $\G = (V, A)$ is the
candidate DAG, $\G_0 = (V, A_0)$ the prior DAG, $N = |V|$, and $R_0 = R(\G_0)$
the interventional reachability set of $\G_0$, the ordered pairs joined by a
directed path in $\G_0$. Their natural upper bounds are
\begin{align*}
  &\SHD(\G, \G_0) \leqslant N(N - 1)/2& &\text{and}&
  &\SID(\G, \G_0) \leqslant N(N - 1),
\end{align*}
since SHD is defined over the $\{vi, v_j\}$ and SID is defined over the
$(v_i, v_j)$. However, these bounds are loose, especially for sparse DAGs. In
the case of SHD,
\begin{equation*}
  \wSHD(\G, \G_0) = \SHD(\G, \G_0) / |A_0|
\end{equation*}
is commonly used in simulation studies comprising ground truth DAGs with very
different sizes and densities \citep[see, for instance, ][]{pgm16}. It can be
interpreted as a structural error rate adjusted by the density of $\G_0$. Its
range depends only on how dense $\G$ is compared with $\G_0$, as the following
theorem shows.

\begin{theorem}
  Let $\G = (V, A)$ and $\G_0 = (V, A_0)$ be two DAGs over the same nodes, with
  $|A_0| > 0$. Then
  \begin{equation*}
    0 \leqslant \wSHD(\G, \G_0) \leqslant 1 + \frac{|A|}{|A_0|},
  \end{equation*}
  where the lower bound is attained if and only if $\G = \G_0$, and the upper
  bound if and only if no node pair is adjacent in both $\G$ and $\G_0$.
  Hence, if $|A| \leqslant c |A_0|$ for some constant $c \geqslant 0$, then
  $\wSHD(\G, \G_0) \leqslant 1 + c$.
\label{theo:bsparse}
\end{theorem}
\begin{proof}[Proof of Theorem~\ref{theo:bsparse}]
  Each unordered node pair $\{v_i, v_j\}$ contributes at most $1$ to SHD, and
  only if it is adjacent in at least one of the two DAGs. Let $s$ be the number
  of pairs that are adjacent in both DAGs with the same direction, and $t$ the
  number of those adjacent in both with opposite directions. The pairs adjacent
  in only one of the two DAGs number $|A| + |A_0| - 2(s + t)$ and each
  contribute $1$; the $t$ reversed pairs also contribute $1$; and the $s$ shared
  arcs contribute $0$. Hence,
  \begin{equation*}
    \SHD(\G, \G_0) = |A| + |A_0| - 2s - t \leqslant |A| + |A_0|,
  \end{equation*}
  with equality if and only if $s = t = 0$, that is, if no node pair is adjacent
  in both DAGs. Dividing by $|A_0| > 0$ gives the upper bound. The lower bound
  holds because SHD is a count, and $\SHD(\G, \G_0) = 0$ if and only if every
  node pair takes the same state in both DAGs, that is, if $\G = \G_0$. The
  last claim follows by substituting $|A| \leqslant c|A_0|$.
\end{proof}

The bound holds for every $N$, and does not depend on it: as long as
$\G$ and $\G_0$ are similarly dense ($c \approx 1$), $\wSHD(\G, \G_0)$ ranges
from $0$ to a small upper bound, which explains its empirical behaviour as
observed in the literature. This is the case, in particular, when both DAGs are
sparse: if $|A| \leqslant c'N$ and $|A_0| \geqslant aN$ for constants $a, c' >
0$, then $c = c'/a$. Some condition on $|A| / |A_0|$ is necessary, because the
upper bound is attained: if $\G_0$ has a single arc and $\G$ has $c'N$ arcs
between other node pairs, then $\wSHD(\G, \G_0) = 1 + c'N$ grows linearly in
$N$. The bound does not hold uniformly over all DAGs, whose number of arcs can
be as large as $N(N - 1)/2$.

Dividing SID by its upper bound gives $\wSID(\G, \G_0) = \SID(\G, \G_0) /
\left(N (N - 1)\right) \in [0, 1]$. Unlike for SHD, no tighter normalisation is
practical: neither the number of arcs nor other simple structural features of
$\G_0$ predict the value of SID, even assuming sparsity. Consider two prior DAGs
with the same number of arcs, $|A_0| = N - 1$, and let $\G$ be the empty DAG. If
$\G_0$ is the chain $v_1 \rarr v_2 \rarr \ldots \rarr v_N$, the empty parent
sets of $\G$ fail to identify the effect of $v_i$ on $v_j$ only when $v_j$ is an
ancestor of $v_i$, so $\SID(\G, \G_0) = N(N - 1)/2$. If $\G_0$ is the star in
which $v_1$ is the only parent of all the other nodes, the empty parent sets do
not adjust for the confounder $v_1$ for any of the ordered pairs of its
children, and do not identify the (null) effect of any child on $v_1$, so
$\SID(\G, \G_0) = (N - 1)^2$. Dividing by $|A_0|$ gives $N/2$ and $N - 1$, both
of which grow with $N$. Dividing by the size of the interventional reachability
set $R_0$ does not work any better: $|R_0| = N(N - 1)/2$ for the chain but
$|R_0| = N - 1$ for the star, giving $1$ and $N - 1$. Only $N(N - 1)$ keeps both
in $[0, 1]$, at $1/2$ and $(N - 1)/N$.

\subsection{Simulation Study}
\label{app:rescaling:sims}

\begin{table}[t]
\begin{center}
  \begin{tabular}{lrrrclrrr}
  \cline{1-4} \cline{6-9}
  \multicolumn{4}{c}{Discrete networks} & &
    \multicolumn{4}{c}{Gaussian networks} \\
  \cline{1-4} \cline{6-9}
  network    & $N$   & $|A_0|$ & $|\Theta|$ & &
    network    & $N$   & $|A_0|$ & $|\Theta|$ \\
  \cline{1-4} \cline{6-9}
  ALARM      & $37$  & $46$    & $509$      & &
    ARTH150    & $107$ & $150$   & $364$      \\
  ANDES      & $223$ & $338$   & $1157$     & &
    ECOLI70    & $46$  & $70$    & $162$      \\
  HAILFINDER & $56$  & $66$    & $2656$     & &
    MAGIC-IRRI & $64$  & $102$   & $230$      \\
  HEPAR2     & $70$  & $123$   & $1453$     & &
    MAGIC-NIAB & $44$  & $66$    & $154$      \\
  INSURANCE  & $27$  & $52$    & $1008$     \\
  LINK       & $724$ & $1125$  & $14211$    \\
  PATHFINDER & $109$ & $195$   & $72079$    \\
  PIGS       & $441$ & $592$   & $5618$     \\
  \cline{1-4} \cline{6-9}
  \end{tabular}
  \caption{Reference BNs from the BN repository \citep{bnrepo} with the
    respective numbers of nodes ($N$), arcs ($|A_0|$) and parameters
    ($|\Theta|$). The Gaussian BNs are the same as in
    Table~\ref{tab:gaussian}.}
  \label{tab:networks}
\end{center}
\end{table}

We verified the empirical behaviour of $\wSHD(\G, \G_0)$, $\wSID(\G, \G_0)$
across data types, learning algorithms and sample sizes with the following
simulation study.
\begin{enumerate}
  \item We selected eight reference discrete BNs and all the available Gaussian
    BNs from the Bayesian network repository \citep{bnrepo}, shown in
    Table~\ref{tab:networks}.
   \item For each network, we generated 20 samples at each of the sample sizes
     $n|\Theta|$, $n = 0.1, 0.2, 0.5, 1, 2, 5$. Values $n < 1$ represent
     small-$n$, large-$|\Theta|$ settings, $n \geqslant 1$ medium and large
     sample sizes.
   \item For each sample, we learned the DAG with two standard structure
     learning algorithms:
     \begin{itemize}
       \item Tabu search \citep{glover}, a greedy search maximising a \linebreak
         goodness-of-fit score using single-arc additions, deletions and
         reversals, scored with the Bayesian information criterion
         \citep[BIC;][]{bic} for both discrete and Gaussian BNs.
       \item The stable variant of the PC algorithm \citep{colombo}, a
         constraint-based algorithm that uses conditional independence tests to
         establish the absence, presence and direction of candidate arcs, at a
         significance threshold of $0.05$: Pearson's $X^2$ test
         \citep{fienberg} for discrete BNs, the exact $t$-test for
         Pearson's correlation \citep{hotelling} for Gaussian ones.
     \end{itemize}
   \item For each learned DAG $\G$, we computed $\SHD(\G, \G_0)$ and
     $\SID(\G, \G_0)$ from the corresponding $\G_0$ in the repository.
\end{enumerate}

Figure~\ref{fig:by-nodes} plots $\wSID(\G, \G_0)$ against $\wSHD(\G, \G_0)$ for
all combinations of learning algorithm, reference BN and $n / |\Theta|$. For
discrete networks, and for Gaussian networks learned with PC, the rescaled
divergences align well along the diagonal or the first quadrant and are bounded
in the square $[0, 1.5] \times [0, 1.5]$. For tabu search and Gaussian networks,
however, $\wSID(\G, \G_0)$ has a markedly different range from
$\wSHD(\G, \G_0)$: the DAGs learned from samples with $n < 1$ have little
overlap with $\G_0$ and more than twice as many arcs as it has, so $|A|$ and
$|A_0|$ are too dissimilar for the rescaling to keep $\wSHD(\G, \G_0)$ in a
small range, as predicted by Theorem~\ref{theo:bsparse}.

\begin{figure}[p]
  \includegraphics[width=\linewidth]{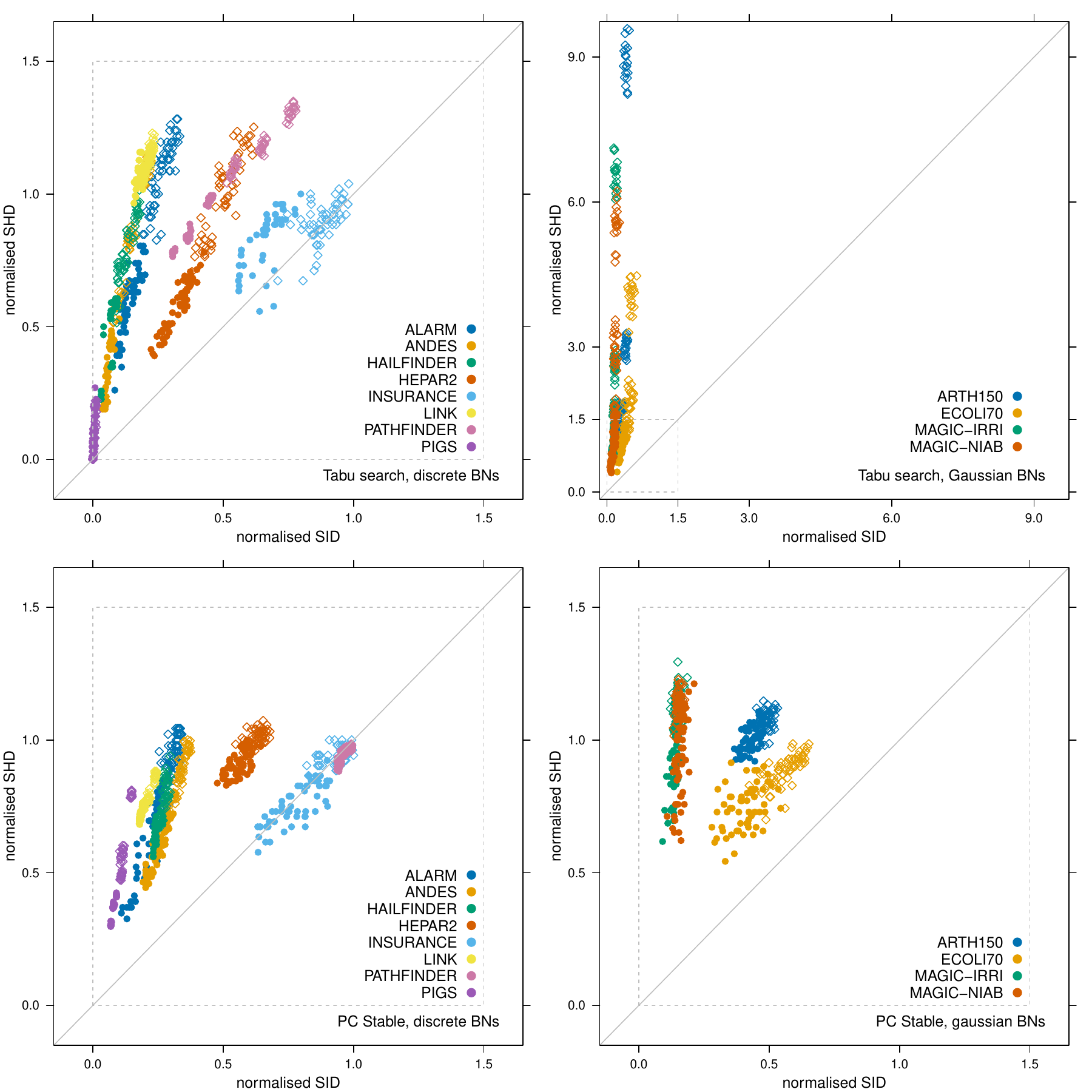}
  \caption{$\wSID(\G, \G_0)$ versus $\wSHD(\G, \G_0)$ for the DAGs learned
    with tabu search (top) and PC-stable (bottom) from the discrete (left) and
    the Gaussian (right) networks of Table~\ref{tab:networks}. DAGs learned
    from samples where $n < 1$ are shown as $\diamond$, those learned from
    larger samples where $n \geqslant 1$ are shown as $\bullet$.}
  \label{fig:by-nodes}
\end{figure}

\clearpage

\section{The Normalising Constant and the Calibration of the Decay}
\label{app:partition}

\subsection{Bounds on the Normalising Constant}
\label{app:partition:bounds}

Recall from Section~\ref{sec:lge} that the prior in \mref{eq:lge} is only
defined up to the normalising constant
\begin{equation*}
  Z(\beta, w) = \sum_{\G \in \mathbb{D}_N}
    \exp\left\{ -\beta E(\G, \G_0; w) \right\},
\end{equation*}
where $\mathbb{D}_N$ is the set of all DAGs over $V$ and
$E(\G, \G_0; w) = w \wSHD(\G, \G_0) + (1 - w) \wSID(\G, \G_0)$ is the energy
\mref{eq:energy} of a candidate DAG $\G$ with respect to the prior DAG $\G_0 =
(V, A_0)$, with $\wSHD = \SHD / |A_0|$ and $\wSID = \SID / \{N(N - 1)\}$. The
robustness results of Section~\ref{sec:robustness} and the absolute rule of
Appendix~\ref{app:absolute} need bounds on it, which we derive by relaxing the
acyclicity constraint as in \citet{imoto04}: SHD decomposes over node pairs,
and the properties of SID supply the matching lower bounds.

\begin{theorem}
  Let $\G_0 = (V, A_0)$ be a non-empty DAG with $N$ nodes, \mbox{$P = N(N -
  1)/2$} node pairs, and write $q = \exp\left\{-\beta w / |A_0|\right\}$.
  Then, for all $\beta \geqslant 0$ and $w \in [0, 1]$,
  \begin{equation*}
    \max\left\{
      (1 + q)^{P - |A_0|}, \; e^{-\beta(1 - w)} (1 + q)^{P}
    \right\}
    \leqslant Z(\beta, w) \leqslant (1 + 2q)^{P}.
  \end{equation*}
\label{theo:partition}
\end{theorem}
\begin{proof}[Proof of Theorem~\ref{theo:partition}]
  The upper bound and the two lower bounds are proved in turn.

  \emph{The product upper bound.} Since $\wSID(\G, \G_0) \geqslant 0$
  and $w \in [0, 1]$,
  \begin{multline*}
    \exp\{-\beta(1 - w) \wSID(\G, \G_0)\} \leqslant 1 \Rightarrow \\
    Z(\beta, w) \leqslant
      \sum_{\G \in \mathbb{D}_N} \exp\left\{ -\beta w \wSHD(\G, \G_0) \right\} =
      \sum_{\G \in \mathbb{D}_N} q^{\SHD(\G, \G_0)}.
  \end{multline*}
  Every DAG is one of the $3^P$ state assignments to the node pairs and all the
  summands are positive, so dropping acyclicity and extending the sum to all
  such assignments can only increase it. Similarly, the extended sum factorises
  over pairs. The factor for each pair is the sum over its three states, which
  is $1$ for the state taken by $\G_0$ and $q$ for each of the other two:
  \begin{equation*}
    Z(\beta, w) \leqslant \prod_{\{v_i, v_j\}} (1 + 2q) = (1 + 2q)^{P}.
  \end{equation*}

  \emph{The supergraph lower bound.} Fix a topological order $\sigma_0$ of
  $\G_0$ and let $F$ be the set of the node pairs that are $\sigma_0$-forward
  and not adjacent in $\G_0$, so that $|F| = P - |A_0|$. For each $S \subseteq
  F$, let $\G_S$ be $\G_0$ with the pairs in $S$ added as $\sigma_0$-forward
  arcs. Every $\G_S$ is consistent with $\sigma_0$ and is therefore a DAG, the
  map $S \mapsto \G_S$ is injective, and $\G_0 \subseteq \G_S$, so
  $\SID(\G_S, \G_0) = 0$ by Proposition 7 in \citet{peters}. Since
  $\SHD(\G_S, \G_0) = |S|$ and all the summands of $Z(\beta, w)$ are positive,
  restricting the sum to $\{\G_S : S \subseteq F\}$ gives
  \begin{equation*}
    Z(\beta, w) \geqslant \sum_{S \subseteq F} q^{|S|} =
      \sum_{k = 0}^{P - |A_0|} \binom{P - |A_0|}{k} q^{k} =
      (1 + q)^{P - |A_0|}.
  \end{equation*}

  \emph{The single-order lower bound.} Fix again a topological order $\sigma_0$
  of $\G_0$ and restrict the sum to the DAGs consistent with $\sigma_0$. Such a
  DAG takes only two of the three states at each node pair, independently
  across pairs: an arc pointing $\sigma_0$-forward, or no arc. Since
  $\sigma_0$ reverses no arc of $\G_0$, exactly one of the two states differs
  from that of $\G_0$ at every pair, so the sum of $q^{\SHD(\G, \G_0)}$ over
  this family factorises as $(1 + q)^P$. On the same family, $\wSID(\G, \G_0)
  \leqslant 1$ gives $\exp\{-\beta(1 - w) \wSID(\G, \G_0)\} \geqslant
  e^{-\beta(1 - w)}$, and therefore
  \begin{equation*}
    Z(\beta, w) \geqslant e^{-\beta(1 - w)}
      \sum_{\G \,\sim\, \sigma_0} q^{\SHD(\G, \G_0)} =
      e^{-\beta(1 - w)} (1 + q)^{P}. \qedhere
  \end{equation*}
\end{proof}

Neither lower bound dominates the other in general. The bounds tighten as
$\beta$ grows: $q \to 0$ and $(1 + 2q)^{P} \to 1$, $(1 + q)^{P - |A_0|} \to 1$,
with a log-scale gap of order $(P + |A_0|) q \to 0$. In contrast, the bounds
become increasingly wide as $\beta \to 0$. We confirm this and
Theorem~\ref{theo:partition} by exhaustive enumeration in
Appendix~\ref{app:enumeration}.

\subsection{Exhaustive Enumeration Study}
\label{app:enumeration}

The bounds from Theorem~\ref{theo:partition} are only useful if they are tight,
which depends on how much relaxing acyclicity overcounts in the upper bound and
on how many DAGs in $\mathbb{D}_N$ the lower bounds leave out. To assess this,
we computed $Z(\beta, w)$ exactly by enumerating all the labelled DAGs over $N =
3, \ldots, 7$ nodes. We considered six prior DAGs spanning a range of shapes at
each $N$, listed in Table~\ref{tab:bracketwidth}: a chain, a star, a reversed
star, a balanced binary tree, a banded DAG in which every node takes as parents
the three that immediately precede it, and the complete DAG. The bounds depend
on $\G_0$ only through $|A_0|$, so the four sparse DAGs, which all have
$|A_0| = N - 1$, share the same bracket and are pooled in the table: at $N =
6$, $w = 0.5$ and $\beta = 10$, their exact $\log Z(\beta, w)$ ranges from
$5.66$ to $6.99$ inside the shared bracket $[3.13, 8.27]$. We evaluated every
prior DAG at $\beta \in \{2, 10, 50, 200\}$ and $w \in \{0.5, 0.75, 1\}$, giving
$72$ configurations for each $N$. The bounds hold in all $360$ configurations.

\begin{table}[t]
\begin{center}
  \small
  \begin{tabular}{lrrrrrcrrrr}
  \cline{3-6} \cline{8-11}
  & & \multicolumn{4}{c}{$w = 0.5$} & & \multicolumn{4}{c}{$w = 1$} \\
  \cline{3-6} \cline{8-11}
  $N$, prior DAG & $|A_0|$ & $\beta = 2$ & $10$ & $50$ & $200$ & &
    $\beta = 2$ & $10$ & $50$ & $200$ \\
  \cline{1-6} \cline{8-11}
  $5$, sparse   & $4$ & $4.63$ & $3.02$ & $0.03$ & $0.00$ & &
    $3.20$ & $0.73$ & $0.00$ & $0.00$ \\
  $5$, banded   & $9$ & $4.87$ & $7.19$ & $1.11$ & $0.00$ & &
    $3.68$ & $2.21$ & $0.04$ & $0.00$ \\
  $5$, complete & $10$ & $4.89$ & $7.94$ & $1.52$ & $0.00$ & &
    $3.72$ & $2.38$ & $0.07$ & $0.00$ \\
  \cline{1-6} \cline{8-11}
  $6$, sparse   & $5$ & $6.58$ & $5.14$ & $0.13$ & $0.00$ & &
    $5.06$ & $1.69$ & $0.00$ & $0.00$ \\
  $6$, banded   & $12$ & $6.87$ & $10.02$ & $2.98$ & $0.01$ & &
    $5.66$ & $3.97$ & $0.23$ & $0.00$ \\
  $6$, complete & $15$ & $6.91$ & $10.23$ & $4.81$ & $0.04$ & &
    $5.75$ & $4.38$ & $0.51$ & $0.00$ \\
  \cline{1-6} \cline{8-11}
  $7$, sparse   & $6$ & $8.92$ & $7.72$ & $0.41$ & $0.00$ & &
    $7.33$ & $3.10$ & $0.01$ & $0.00$ \\
  $7$, banded   & $15$ & $9.28$ & $12.33$ & $5.69$ & $0.05$ & &
    $8.04$ & $6.13$ & $0.71$ & $0.00$ \\
  $7$, complete & $21$ & $9.35$ & $12.67$ & $9.98$ & $0.36$ & &
    $8.18$ & $6.81$ & $1.71$ & $0.00$ \\
  \cline{1-6} \cline{8-11}
  \end{tabular}
  \caption{Numbers of arcs $|A_0|$ of the prior DAGs used in the enumeration
    study, and width of the bracket of Theorem~\ref{theo:partition} on the log
    scale, for $N = 5, 6, 7$. ``Sparse'' collects the chain, the star, the
    reversed star and the tree, which all have $|A_0| = N - 1$.}
  \label{tab:bracketwidth}
\end{center}
\end{table}

Table~\ref{tab:bracketwidth} reports the width of the bracket on the log
scale, that is, the logarithm of the upper bound minus the logarithm of the
larger of the two lower bounds. Two features stand out.
\begin{itemize}
  \item \emph{The width peaks where the two lower bounds cross.} For small
    $\beta$, the single-order bound $e^{-\beta(1 - w)} (1 + q)^P$ is the
    larger of the two, and its $\beta(1 - w)$ term keeps the bracket wide. For
    large $\beta$, the supergraph bound $(1 + q)^{P - |A_0|}$ takes over, and
    the width becomes $P \log(1 + 2q) - (P - |A_0|) \log(1 + q) \approx
    (P + |A_0|) q$, which vanishes as $q \to 0$. The crossing moves to larger
    $\beta$ in proportion to $|A_0|$, because $q$ depends on $\beta$ only
    through $\beta w / |A_0|$: at $N = 6$ and $w = 0.5$, it falls before
    $\beta = 10$ for the sparse prior DAGs and at $\beta = 10$ for the
    complete one. At $w = 1$, the $\beta(1 - w)$ term vanishes, the
    single-order bound is always the larger, and the width decreases
    monotonically in $\beta$.
  \item \emph{The width grows with $N$ and with smaller $w$, but not faster
    than $\log Z(\beta, w)$.} A smaller $w$ widens the bracket through a
    larger $q$ and a larger $\beta(1 - w)$ term. At fixed $\beta$ and $w$, the
    width grows with $N$, but $\log Z(\beta, w)$ grows faster: for the chain
    at $\beta = 2$ and $w = 0.5$, the ratio between the two falls from
    $0.562$ at $N = 5$ to $0.490$ at $N = 7$.
\end{itemize}

\subsection{Absolute Calibration of the Decay}
\label{app:absolute}

The rule \mref{eq:ballrule} in Section~\ref{sec:calibration} controls the mass
of the first shell $S_1(\G_0)$ relative to that of $\G_0$, and needs no
normalising constant. Here, we relate it to the absolute prior probabilities of
$S_1(\G_0)$ and $\G_0$, which require the bounds on $Z(\beta, w)$ from
Theorem~\ref{theo:partition}. Recall that $q = \exp\left\{-\beta w /
|A_0|\right\}$, that $P = N(N - 1)/2$, that $S_k(\G_0)$ is the shell of the
DAGs at SHD $k$ from $\G_0$, that $r$ is the mass ratio the shell rule allows
to $S_1(\G_0)$, and that $\bar{d} = |A_0| / N$ is the average in-degree of
$\G_0$. A DAG is determined by the states of its $P$ node pairs and SHD counts
those at which two DAGs disagree, so choosing $k$ pairs and one of the two
other states at each enumerates a superset of $S_k(\G_0)$:
\begin{equation}
  \left| S_k(\G_0) \right| \leqslant \binom{P}{k} 2^k,
\label{eq:shellbound}
\end{equation}
because not all the graphs so obtained are acyclic.

We first relate \mref{eq:ballrule} to the share of the whole prior that the
first shell carries. Partition $Z(\beta, w) = 1 + M + T$ into the unnormalised
prior of $\G_0$, which is $1$, the mass $M$ of $S_1(\G_0)$ and the mass $T$ of
the DAGs $\G : \SHD(\G, \G_0) \geqslant 2$. The shell rule imposes
$M \leqslant r$, and $T \geqslant 0$, so $\Prob(S_1(\G_0)) = M / (1 + M + T)
\leqslant r / (1 + r)$: requiring $S_1(\G_0)$ to be no heavier than $\G_0$
itself, $r = 1$, forces it to hold at most half of the prior mass.

This does not bound $\Prob(\G_0) = 1 / (1 + M + T)$: that requires bounding
$T$, which in turn requires $Z(\beta, w)$. The upper bound of
Theorem~\ref{theo:partition} gives $\Prob(\G_0)
= 1 / Z(\beta, w) \geqslant (1 + 2q)^{-P} \geqslant \exp\{-N(N - 1)q\}$,
leading to the second rule:\footnote{It is conservative, since
$\exp(-\epsilon) \geqslant 1 - \epsilon$.}
\begin{equation}
  \Prob(\G_0) \geqslant 1 - \epsilon
  \qquad\text{whenever}\qquad
  \beta \geqslant \frac{|A_0|}{w} \log \frac{N(N - 1)}{\epsilon}.
\label{eq:absrule}
\end{equation}

The two rules \mref{eq:ballrule} and \mref{eq:absrule} have very similar
expressions. The following lemma shows that they are the first term and the
whole sum of a series over shells with increasing $k$.

\begin{lemma}
  Let $q = \exp\left\{-\beta w / |A_0|\right\}$ and $\lambda = N(N - 1) q$.
  Then
  \begin{equation*}
    \frac{\Prob(S_k(\G_0))}{\Prob(\G_0)} \leqslant
      \left| S_k(\G_0) \right| q^k \leqslant \binom{P}{k} (2q)^k \leqslant
      \frac{\lambda^k}{k!}
    \qquad\text{for every } k,
  \end{equation*}
  and summing the middle term over $k$ returns the product bound $(1 + 2q)^P$ of
  Theorem~\ref{theo:partition}.
\label{theo:shells}
\end{lemma}

\begin{proof}[Proof of Lemma~\ref{theo:shells}]
  By \mref{eq:shellbound}, $|S_k(\G_0)| q^k \leqslant \binom{P}{k} 2^k q^k
  \leqslant (2Pq)^k / k! = \lambda^k / k!$, because $\binom{P}{k} \leqslant
  P^k / k!$. Since $\wSID \geqslant 0$,
  \begin{equation*}
    E(\G, \G_0; w) \geqslant w \wSHD(\G, \G_0) = \frac{wk}{|A_0|}
    \qquad\text{for every } \G \in S_k(\G_0),
  \end{equation*}
  so each such $\G$ has unnormalised prior at most $q^k$, as in the proof of
  Theorem~\ref{theo:partition}, while $\G_0$ has unnormalised prior $e^{0} = 1$.
  Summing over $S_k(\G_0)$ and dividing gives the first inequality. Summing the
  middle term over $k = 0, \ldots, P$ gives
  $\sum_k \binom{P}{k} (2q)^k = (1 + 2q)^P$ by the binomial theorem, which is
  the product bound of Theorem~\ref{theo:partition}; the shells therefore
  exhaust $Z(\beta, w)$, and the rules \mref{eq:ballrule} and
  \mref{eq:absrule} are its truncations.
\end{proof}

Lemma~\ref{theo:shells} bounds the mass of each shell by the number of graphs
it could contain, $\binom{P}{k} 2^k$, which ignores both acyclicity and SID. To
compare the calibration thresholds themselves, we need bounds that hold for the
exact masses of the shells. Write $\beta_k$ for the smallest $\beta$ at which
the first $k$ shells carry at most $r$ times the prior probability of $\G_0$,
\begin{equation*}
  \sum_{j = 1}^{k} \frac{\Prob(S_j(\G_0))}{\Prob(\G_0)} \leqslant r,
\end{equation*}
and $\beta_*$ for the same over all the shells, that is,
$Z(\beta, w) - 1 \leqslant r$. Furthermore, let
\begin{align*}
  &n_+ = N(N - 1) - |R_0| - |A_0|&
  &\text{and}&
  &\kappa = (1 + r)^{N(N - 1) / n_+}.
\end{align*}
where $R_0 = R(\G_0)$ is the interventional reachability set of $\G_0$ from
Section~\ref{sec:notation}. $n_+$ is the number of arcs that can be added to
$\G_0$ without creating a cycle (Appendix~\ref{app:calibration}), so that
$|S_1(\G_0)| = n_+ + |A_0| + \#\{\text{acyclic reversals}\}$; it is positive
unless $\G_0$ is complete, since a non-adjacent pair joined by a directed path
still accepts the arc in the direction of the path.

\begin{proposition}
  If $0 < r < n_+$, then $\beta_1 \leqslant \beta_2 \leqslant \ldots
  \leqslant \beta_*$,
  \begin{align*}
    &\frac{|A_0|}{w} \log \frac{n_+}{r} \leqslant \beta_1 \leqslant
      \frac{|A_0|}{w} \log \frac{|S_1(\G_0)|}{r},&
    &\beta_* \leqslant
      \frac{|A_0|}{w} \log \frac{n_+ + 2\kappa |A_0|}{\log(1 + r)},
  \end{align*}
  and therefore
  \begin{equation*}
    0 \leqslant \beta_* - \beta_1 \leqslant \frac{|A_0|}{w} \left[
      \log \frac{r}{\log(1 + r)} +
      \log\left(1 + \frac{2\kappa |A_0|}{n_+}\right) \right].
  \end{equation*}
\label{theo:thresholds}
\end{proposition}

\begin{proof}
  Each partial sum of the shell masses decreases in $\beta$ and increases in
  $k$, so the thresholds are well defined and increase with $k$. For $\beta_1$,
  the $n_+$ additions in $S_1(\G_0)$ have $\SID(\G, \G_0) = 0$ and therefore
  energy exactly $w / |A_0|$, while all the DAGs in $S_1(\G_0)$ have energy at
  least $w / |A_0|$. Hence, $n_+ q \leqslant \Prob(S_1(\G_0)) / \Prob(\G_0)
  \leqslant |S_1(\G_0)| q$, and solving $n_+ q = r$ and $|S_1(\G_0)| q = r$
  for $\beta$ gives the bracket on $\beta_1$.

  For $\beta_*$, every $\G \neq \G_0$ contributes at most $q^{\SHD(\G, \G_0)}$
  to $Z(\beta, w) - 1$, since $\wSID \geqslant 0$. Consider first the DAGs that
  contain every arc of $\G_0$. A node pair that is not adjacent in $\G_0$ but is
  joined by a directed path in it can only be left empty or take the arc in the
  direction of the path, since the other would close a cycle; there are
  $|R_0| - |A_0|$ such pairs. Relaxing acyclicity for the other
  $P - |R_0|$ non-adjacent pairs, these DAGs contribute at most
  \begin{equation*}
    (1 + 2q)^{P - |R_0|} (1 + q)^{|R_0| - |A_0|} - 1 \leqslant e^{n_+ q} - 1.
  \end{equation*}
  The remaining DAGs differ from $\G_0$ in at least one of its arcs, and
  relaxing acyclicity for all the pairs they contribute at most
  \begin{equation*}
    (1 + 2q)^{P - |A_0|} \left\{ (1 + 2q)^{|A_0|} - 1 \right\} \leqslant
    e^{2(P - |A_0|)q} \left( e^{2|A_0|q} - 1 \right) \leqslant
    2|A_0| q \, e^{N(N - 1) q}.
  \end{equation*}
  For $q \leqslant \log(1 + r) / n_+$, $e^{N(N - 1)q} \leqslant \kappa$, and
  since $e^a - 1 + b \leqslant e^{a + b} - 1$ for $a, b \geqslant 0$,
  \begin{equation*}
    Z(\beta, w) - 1 \leqslant e^{n_+ q} - 1 + 2\kappa |A_0| q \leqslant
      \exp\left\{ (n_+ + 2\kappa |A_0|) q \right\} - 1.
  \end{equation*}
  The right-hand side equals $r$ at $q = \log(1 + r) / (n_+ + 2\kappa |A_0|)$,
  which is smaller than $\log(1 + r) / n_+$, so $Z(\beta, w) - 1 \leqslant r$
  for every $\beta$ at which $q$ is at most that value. This gives the bound on
  $\beta_*$. Subtracting the lower bound on $\beta_1$ gives the bound on
  $\beta_* - \beta_1$, which is non-negative because $\beta_1 \leqslant
  \beta_*$.
\end{proof}

Neither bracketed term of the bound on $\beta_* - \beta_1$ grows with $N$. The
first is a constant, $0.37$ at $r = 1$. As for the second, every node pair
contributes at most one ordered pair to $R_0$, so $|R_0| \leqslant N(N - 1)/2$
and $n_+ \geqslant N(N - 1)/2 - |A_0|$: for a sparse $\G_0$ with $|A_0| =
\bar{d} N$, $\kappa$ is bounded and $|A_0| / n_+ = O(\bar{d} / N)$, so the
second term vanishes as $N$ grows. On the other hand, $\beta_1 \geqslant
(|A_0|/w) \log(n_+ / r)$ grows with $\log N$. Hence, everything outside
$S_1(\G_0)$ impacts $O(1 / \log N)$ of the calibration: the first shell is a
vanishing fraction of $\mathbb{D}_N$ but largely controls the calibration, and
the shell rule \mref{eq:ballrule} is the upper end of the bracket on $\beta_1$.
On the reference networks of Appendix~\ref{app:numerical}, the bracket on
$\beta_1$ is less than $1\%$ wide, and the bound on $\beta_* - \beta_1$ is
between $4.6\%$ and $7.1\%$ of $\beta_1$. Bounding $\beta_*$ by solving for $q$
numerically the sum of the two bounds in the proof, before relaxing them into
exponentials, it exceeds the shell rule by only $4.0\%$ to $5.0\%$. We also
checked Proposition~\ref{theo:thresholds} against the exact thresholds computed
from the enumeration of Appendix~\ref{app:enumeration}, for the five prior DAGs
with $n_+ > 0$ at $N = 5, 6, 7$, $w \in \{0.5, 0.75, 1\}$ and $r \in \{0.5, 1,
2, 5\}$: the bounds hold in all $180$ settings, and $\beta_* - \beta_1$ is at
most $0.34$ times its bound.

\subsection{The First Two Shells}
\label{app:calibration}

The closed form \mref{eq:ballsize} for the size of the first shell follows
from counting the arc edits that keep $\G_0$ acyclic.

\begin{lemma}
  Let $\G_0 = (V, A_0)$ be a DAG with $N$ nodes and interventional reachability
  set $R_0$, and call an arc of $\G_0$ an \emph{acyclic reversal} if reversing
  it does not create a cycle. Then
  \begin{equation*}
    \left| S_1(\G_0) \right| =
      N(N - 1) - |R_0| + \#\{\text{acyclic reversals}\}.
  \end{equation*}
\label{theo:ballsize}
\end{lemma}
\begin{proof}
  The first shell collects the DAGs obtained from $\G_0$ by a single addition,
  deletion or reversal, and we count each kind of edit in turn. Deleting any
  arc of $\G_0$ leaves a DAG, so the deletions number $|A_0|$. A non-adjacent
  node pair accepts an arc in either direction unless one of them introduces a
  cycle, which happens exactly when the pair is joined by a directed path in
  $\G_0$: such pairs are counted by $R_0$ but not by $A_0$, so they number
  $|R_0| - |A_0|$, and each accepts only the arc in the direction of the path.
  Therefore, the acyclic additions number
  \begin{equation*}
    2\left( \frac{N(N - 1)}{2} - |A_0| \right) - \left( |R_0| - |A_0| \right)
      = N(N - 1) - |R_0| - |A_0|,
  \end{equation*}
  which is the $n_+$ of Appendix~\ref{app:absolute}. Adding the $|A_0|$
  deletions and the acyclic reversals gives the claim.
\end{proof}

The second shell has no comparably simple closed form, because the two edits no
longer act independently. Writing $\G \cup F$ and $\G \setminus F$ for the
directed graphs with arc sets $A \cup F$ and $A \setminus F$, every DAG at
$\SHD = 2$ from $\G_0$ is of the form $(\G_0 \setminus D) \cup F$, with $D$ the
arcs removed from $\G_0$ and $F$ the arcs added to it, and whether it is
acyclic depends on the directed paths of $\G_0 \setminus D$ between the
endpoints of the arcs in $F$.
However, $|S_2(\G_0)|$ can still be computed exactly in time polynomial in $N$,
by enumerating the pairs of edits and testing each for acyclicity on the
transitive closure of $\G_0 \setminus D$. We did so for the four networks of
Table~\ref{tab:calibration}, and checked the computation against the exhaustive
enumerations of Appendix~\ref{app:enumeration}, which it reproduces exactly for
all six prior DAGs at $N = 3, \ldots, 7$. Truncating the series of
Lemma~\ref{theo:shells} at $k = 2$ then gives a further rule, the positive root
of $|S_1(\G_0)| q + |S_2(\G_0)| q^2 = r$, which is the threshold $\beta_2$ of
Appendix~\ref{app:absolute}. Acyclicity costs the second shell little on the
sparse prior DAGs that calibration is aimed at: $|S_2(\G_0)|$ is $86\%$ to
$95\%$ of the bound \mref{eq:shellbound} on the four networks, against $43\%$
on the banded and the complete prior DAGs at $N = 7$. On the four networks,
$87\%$ to $95\%$ of the second shell consists of two additions, which have
$\G_0 \subseteq \G$, vanishing SID and energy exactly $2w / |A_0|$: the bound of
Lemma~\ref{theo:shells} is nearly attained for the DAGs we consider in
practical settings.

\subsection{Numerical Evaluation on Reference Networks}
\label{app:numerical}

We computed the calibration rules from Section~\ref{sec:lge} on the four
Gaussian networks of Table~\ref{tab:gaussian}, taking their DAGs as the prior
$\G_0$, to see what values of $\beta$ and $w$ they suggest in practice and how
these values depend on $\G_0$. The same networks are used in the simulations of
Section~\ref{sec:sims}, so these results also motivate the values of $\beta$
and $w$ we used there. Recall that $w^* = 2\bar{d} / (1 + 2\bar{d})$ in
\mref{eq:wstar} is the balance point, with $\bar{d} = |A_0| / N$ the average
in-degree of $\G_0$; that the shell rule \mref{eq:ballrule} sets $\beta$ from
the mass ratio $r$ allowed to the first shell $S_1(\G_0)$, and the absolute
rule \mref{eq:absrule} from the bound $\epsilon$ on $1 - \Prob(\G_0)$; that
$q = \exp\{-\beta w / |A_0|\}$ and $\lambda = N(N - 1) q$; and that
$\beta^*(m)$ in \mref{eq:betastar} is the robust decay for a $\G_0$ that is
wrong at $m$ node pairs. Section~\ref{sec:experiments} summarises the results
in three points. The first, that $w^*$ varies little between prior DAGs that
are similarly sparse, follows from \mref{eq:wstar}: $w^*$ increases with
$\bar{d}$, because the denser the prior DAG the less a single arc difference
costs in $\wSHD$, but it flattens quickly, from $3/4$ at $\bar{d} = 1.5$ to
$6/7$ at $\bar{d} = 3$ and $10/11$ at $\bar{d} = 5$. All the reference
networks of Table~\ref{tab:networks} have $\bar{d} \in [1.18, 1.93]$, so
$w^* \approx 3/4$ is a reasonable default for sparse prior DAGs. The other
two points need more discussion, which we give in the next two subsections.

\subsection{The Decay for a Correct Prior DAG}
\label{app:numerical:decay}

\begin{table}[b]
\begin{center}
  \small
  \setlength{\tabcolsep}{4pt}
  \begin{tabular}{lrrcrrrcrrr}
  \cline{4-6} \cline{8-10}
  & & & \multicolumn{3}{c}{$w = 0.5$} & & \multicolumn{3}{c}{$w = 0.75$} \\
  \cline{4-6} \cline{8-10}
    network & $|S_1(\G_0)|$ & $|S_2(\G_0)|$ &
    $r = 1$ & $\epsilon = 0.1$ & at $500$ & &
    $r = 1$ & $\epsilon = 0.1$ & at $500$ \\
  \cline{1-3} \cline{4-6} \cline{8-10}
  ARTH150    & $11037$ & $60.84 \times 10^6$ & $2793$ & $3492$ & $2050$ & & $1862$ & $2328$ & $896$  \\
  ECOLI70    &  $1919$ &  $1.83 \times 10^6$ & $1058$ & $1391$ & $51$   & & $706$  & $928$  & $9$    \\
  MAGIC-NIAB &  $1798$ &  $1.61 \times 10^6$ & $989$  & $1300$ & $39$   & & $660$  & $867$  & $6$    \\
  MAGIC-IRRI &  $3813$ &  $7.25 \times 10^6$ & $1682$ & $2163$ & $320$  & & $1121$ & $1442$ & $95$   \\
  \cline{1-3} \cline{4-6} \cline{8-10}
  \end{tabular}
  \caption{Calibration constants for the networks of Table~\ref{tab:gaussian}.
    $|S_1(\G_0)|$ and $|S_2(\G_0)|$ are computed exactly as described in
    Appendix~\ref{app:calibration}.
    The columns for $r = 1$ and $\epsilon = 0.1$ give the smallest $\beta$
    satisfying \mref{eq:ballrule}, and \mref{eq:absrule} with $\Prob(\G_0)
    \geqslant 0.9$, which by Lemma~\ref{theo:shells} is the same rule read over
    all the shells at a stricter target. The columns for ``at $500$'' show
    $\Prob(S_1(\G_0)) / \Prob(\G_0)$ at $\beta = 500$, the largest value in
    Section~\ref{sec:sims}, computed from the exact SID of every DAG in the
    shell.}
  \label{tab:calibration}
\end{center}
\end{table}

The $\beta$ suggested by the shell rule \mref{eq:ballrule} and the absolute rule
\mref{eq:absrule} for a correct $\G_0$ grows with its number of arcs, and is
largely determined by the first shell. Unlike $w^*$, it scales with $|A_0| / w$,
and thus differs markedly among the four DAGs as shown in
Table~\ref{tab:calibration}. Both rules only constrain $\beta w$, so the two
halves of the table differ by exactly the ratio of the two weights. Larger
$\G_0$s suggest larger values of $\beta$ because $\wSHD$ increases more slowly
with the number of wrong arcs as $|A_0|$ grows: ARTH150 has $2.3$ times as many
arcs as MAGIC-NIAB, and the suggested $\beta$ is $2.8$ times larger.
Furthermore, larger $\G_0$s have larger $S_1(\G_0)$, and \mref{eq:ballrule}
grows with $\log |S_1(\G_0)|$: this accounts for the remaining factor of
$1.24$. Controlling $S_2(\G_0)$ as well changes $\beta$ by $4.1\%$ on ECOLI70,
in line with Proposition~\ref{theo:thresholds}. The strongest prior used in the
simulations in Section~\ref{sec:sims}, $\beta w = 1000$, is enough for all
networks but ARTH150, which is in fact the hardest network to learn.

The normalising constant enters the calibration only through \mref{eq:absrule},
and the bounds of Theorem~\ref{theo:partition} depend on $\G_0$ only through
$\lambda = N(N - 1) q$, up to a factor $1 - |A_0| / P$ in the exponent of the
upper bound: they bracket $\Prob(\G_0)$ between $e^{-\lambda}$ and about
$e^{-\lambda / 2}$. What they say is therefore a property of the regime, not of
the network. In the regime of the rules, $\lambda \leqslant 1$, the bracket is
tight: at the $r = 1$ thresholds of Table~\ref{tab:calibration}, $\Prob(\G_0)
\in [0.34, 0.61]$ on all four networks, and at $\epsilon = 0.1$, $\Prob(\G_0)
\in [0.90, 0.95]$. Below it, the bounds on $\log Z(\beta, w)$ are looser but
still informative: on ECOLI70 at $\beta = 50$ the shells grow instead of
decaying, $\lambda = 1448$, and $\Prob(\G_0) \leqslant e^{-524}$. At the other
end, $\beta w = 1000$ puts at least $0.998$ of the prior on $\G_0$ for ECOLI70
and MAGIC-NIAB and between $0.80$ and $0.90$ for MAGIC-IRRI, but at most $0.001$
for ARTH150.

\subsection{The Decay for a Wrong Prior DAG}
\label{app:numerical:robust}

A prior DAG that may be wrong calls for a slower decay than a correct one, and
choosing a decay that is too fast makes the prior exponentially harder for the
data to overturn. The rules of Section~\ref{sec:calibration} and the bound of
Section~\ref{sec:robustness} pull $\beta$ in opposite directions: a larger
$\G_0$ needs a faster decay to concentrate; a more wrong $\G_0$ needs a slower
one to keep mass on $\G_T$. Corrupting $10\%$, $20\%$ and $50\%$ of the arcs of
ECOLI70, as Section~\ref{sec:sims} does, leaves $m = 6.2$, $11.9$ and $26.7$
wrong arcs on average; then \mref{eq:betastar} gives $\beta^*(m) = 619$, $527$
and $429$ at $w = 0.5$, in contrast to the $\beta = 1058$ from
\mref{eq:ballrule}. Hence, assuming that $\G_0$ is correct when $10\%$ of its
arcs are wrong makes the decay $1.7$ times faster than it should be. The $\beta
= 500$ in the grid of Section~\ref{sec:sims} falls within the range of the
robust decays and well below the shell rule. The weight enters through $\beta w$
here too: at the lightest perturbation, $\beta^*(m) w$ rises from $245$ at $w =
0.25$ to $407$ at $w = 1$, because the $\wSID$ term of the energy shrinks as $w$
grows, and stays below the $529$ of the $r = 1$ threshold at every $w$. A wrong
$\G_0$ requires a slower decay than a right one, regardless of $w$. Choosing a
decay that is too fast has a steep price in evidence: by
Corollary~\ref{theo:evidence}, with $10\%$ of the arcs wrong the prior odds of
$\G_0$ against $\G_T$ are $e^{8.9}$ at $\beta = 50$ and $e^{88.8}$ at $\beta =
500$, so at the fast end of the grid the data must favour $\G_T$ about
$10^{34}$ times more strongly than at the slow end. The simulation results from
Section~\ref{sec:sims} agree:
at the larger sample sizes, the best $\beta$ falls from $500$ to $200$ as the
perturbation grows from $20\%$ to $50\%$, which is where $\beta^*(m)$ drops
below $500$.

\clearpage

\section{Proofs from Section~\ref{sec:robustness}}
\label{app:robust}

Throughout this appendix, $\G_T$ is the true DAG, $m = \SHD(\G_T, \G_0)$ the
perturbation level of the prior DAG $\G_0 = (V, A_0)$, $\Pi_{v_i}(\G)$ the
parent set of $v_i$ in $\G$, $R_0 = R(\G_0)$ the set of ordered pairs joined by
a directed path in $\G_0$, $P = N(N - 1) / 2$, and $Z(\beta, w)$ the
normalising constant bounded in Appendix~\ref{app:partition:bounds}. $\PD$,
$\sigma_m$, $q$, $m_w$ and $\kappa_w$ are as in
Section~\ref{sec:robustness}, and $\D$, $p_T$ and $\Delta$ as in
Corollary~\ref{theo:evidence}.

\begin{proof}[Proof of Lemma~\ref{theo:sidperturb}]
  We first show that a node at which $\G$ and $\G_0$ agree does not contribute
  to SID. Suppose that $\Pi_{v_i}(\G) = \Pi_{v_i}(\G_0)$; we check that
  $\Pi_{v_i}(\G_0)$ is a valid adjustment set for $(\G_0, v_i, v_j)$ for every
  $v_j \neq v_i$, in the sense of Section~\ref{sec:notation}.

  If $v_j \in \Pi_{v_i}(\G_0)$, then $v_j$ is a parent of $v_i$ in $\G_0$, so a
  directed path from $v_i$ to $v_j$ would introduce a cycle and
  $(v_i, v_j) \not\in R_0$: the first branch of the definition is
  satisfied. If $v_j \not\in \Pi_{v_i}(\G_0)$, we check the second. No node of
  $\Pi_{v_i}(\G_0)$ is a descendant of a node lying on a directed path from
  $v_i$ to $v_j$, since every such node is $v_i$ or one of its descendants and
  a parent of $v_i$ that were a descendant of $v_i$ would close a cycle. It
  remains to show that $\Pi_{v_i}(\G_0)$ blocks every non-directed path $\pi$
  between $v_i$ and $v_j$. If $\pi$ leaves $v_i$ by an incoming arc
  $v_i \larr p$, then $p \in \Pi_{v_i}(\G_0)$, and $p$ is not a collider on
  $\pi$ because the arc $p \rarr v_i$ points away from it, so conditioning on
  $p$ blocks $\pi$. If instead $\pi$ leaves $v_i$ by an outgoing arc and is not
  a directed path from $v_i$ to $v_j$, it must reverse orientation somewhere,
  and the first vertex $u$ at which it does is a collider on $\pi$.
  That $u$ is reached from $v_i$ by a directed segment, so $u$ and all its
  descendants are descendants of $v_i$ and none of them lies in
  $\Pi_{v_i}(\G_0)$, again by acyclicity. The collider is therefore closed and
  $\pi$ is blocked. Hence, $\Pi_{v_i}(\G_0)$ is valid, and the pair is not
  counted.

  Therefore, only the nodes counted by $\PD(\G, \G_0)$ can contribute, and each
  contributes at most the $N - 1$ pairs it forms with the other nodes, which
  gives \linebreak $\SID(\G, \G_0) \leqslant (N - 1) \PD(\G, \G_0)$. For the
  second bound, the two DAGs differ at exactly $\SHD(\G, \G_0)$ node pairs; and
  a pair $\{v_i, v_j\}$ on which they differ changes the parent set of $v_j$
  alone if the difference is the presence or absence of an arc, and of both
  $v_i$ and $v_j$ if it is an orientation, so at most two nodes per differing
  pair. Then, $\PD \leqslant N$ trivially. Dividing by $N(N - 1)$ gives
  $\wSID(\G, \G_0) \leqslant \PD/N \leqslant \min\{1, 2m/N\}$.
\end{proof}

\begin{proof}[Proof of Theorem~\ref{theo:minimax}]
  For the lower bound, Lemma~\ref{theo:sidperturb} gives
  \begin{equation*}
    E(\G_T, \G_0; w) \leqslant \frac{wm}{|A_0|} + (1 - w)\sigma_m,
  \end{equation*}
  so that the unnormalised prior at $\G_T$ is at least the exponential in
  \mref{eq:robust}. The product
  bound of Theorem~\ref{theo:partition} and $\log(1 + y) \leqslant y$ give
  $Z(\beta, w) \leqslant (1 + 2q)^P \leqslant \exp\{2Pq\} = \exp\{N(N-1)q\}$,
  and dividing one by the other gives \mref{eq:robust}. The bound is uniform
  over the DAGs at $\SHD = m$, so it holds for all of them.

  For the optimisation, reparametrise by $u = \beta w / |A_0| \geqslant 0$.
  Then $\beta(1 - w)\sigma_m = u(1 - w)\sigma_m|A_0|/w$ and the exponent of
  \mref{eq:robust} is $-m_w u - N(N-1)e^{-u}$, which is concave in $u$ and
  maximised where $m_w = N(N-1)e^{-u}$, that is at $u = \log\{N(N-1)/m_w\}$,
  which is non-negative by the proviso $m_w \leqslant N(N - 1)$. Substituting
  back gives \mref{eq:betastar} and the value $m_w + m_w\log\{N(N-1)/m_w\}$
  for the exponent. When $2m \leqslant N$, $\sigma_m = 2m/N$ and
  $m_w = m\{1 + 2\bar{d}(1 - w)/w\} = m\kappa_w$.
\end{proof}

The guarantee in \mref{eq:betastar} is matched by an upper bound on the best
worst-case probability $V(m) = \sup_\beta \min \{ \Prob(\G_T) :
\SHD(\G_T, \G_0) = m \}$ that any $\beta$ can achieve.

\begin{proposition}
  Let $q$, $\sigma_m$ and $m_w$ be as in Theorem~\ref{theo:minimax}.
  For every $\beta$, every DAG $\G_T$ with $\SHD(\G_T, \G_0) = m$
  has prior probability at most $q^m (1 + q)^{-(P - |A_0|)}$. Hence, if
  $m < P - |A_0|$ and $m_w \leqslant N(N - 1)$, $V(m)$ lies between the value
  in \mref{eq:betastar} and the maximum of this bound over $\beta$, which are
  within a factor
  \begin{equation*}
    \left( \frac{2P}{P - |A_0| - m} \right)^m
    \left( \frac{m}{m_w} \right)^m
    \left( \frac{2eP}{m_w} \right)^{m_w - m},
  \end{equation*}
  of each other. The factor reduces to $\left\{2P / (P - |A_0| - m)\right\}^m$
  at $w = 1$, where $m_w = m$.
\label{theo:minimax-upper}
\end{proposition}
\begin{proof}
  Fix $\beta$ and take any $\G_T$ at $\SHD = m$. Its unnormalised prior is at
  most $q^m$ because $\wSID \geqslant 0$, and $Z(\beta, w) \geqslant (1 + q)^{P
  - |A_0|}$ by Theorem~\ref{theo:partition}, so $\Prob(\G_T) \leqslant q^m (1 +
  q)^{-(P - |A_0|)}$ for every such DAG. This holds for every $\beta$, so it
  also bounds the best worst-case value any $\beta$ can achieve. Writing $p = P
  - |A_0| > m$, the maximiser of $q^m (1 + q)^{-p}$ over $q > 0$ is $q = m/(p -
  m)$. If $2m > p$, it exceeds $1$ and cannot be attained with $\beta \geqslant
  0$, but the unconstrained maximum still bounds the constrained one. In both
  cases, the right-hand side is at most \mbox{$[m/(p-m)]^m (1 - m/p)^p \leqslant
  [m/(p-m)]^m e^{-m}$}. The lower bound in \mref{eq:betastar} is
  $[m_w/(2P)]^{m_w} e^{-m_w}$, and the ratio of the two is at most
  \begin{equation*}
    \left( \frac{m}{p - m} \right)^m e^{-m}
    \left( \frac{2P}{m_w} \right)^{m_w} e^{m_w} =
    \left( \frac{2P}{p - m} \right)^m
    \left( \frac{m}{m_w} \right)^m
    \left( \frac{2eP}{m_w} \right)^{m_w - m},
  \end{equation*}
  which is $[2P/(p - m)]^m$ at $w = 1$, since then $m_w = m$.
\end{proof}

\begin{proof}[Proof of Corollary~\ref{theo:evidence}]
  By the identity in Section~\ref{sec:robustness}, and dividing by $n$, the log
  posterior odds equal $n\{\Delta + o_p(1)\} - \beta E(\G_T, \G_0; w)$, which
  diverges to $+\infty$ in probability when $\Delta > 0$ and crosses zero at $n
  \approx \beta E(\G_T, \G_0; w) / \Delta$. Lemma~\ref{theo:sidperturb} bounds
  $E(\G_T, \G_0; w)$ as in Theorem~\ref{theo:minimax}, and substituting
  $\beta^*(m)$ from \mref{eq:betastar} gives the last bound. In the Markov
  equivalent case, the two DAGs represent the same distributions, so a
  score-equivalent score gives them the same marginal likelihood, and the
  identity reduces to the prior odds $\exp\{-\beta E(\G_T, \G_0; w)\} < 1$. The
  energy is positive because $\wSHD(\G_T, \G_0) > 0$ for $\G_T \neq \G_0$.
\end{proof}

Corollary~\ref{theo:evidence} does not cover the case in which $\G_0$ contains
every arc of $\G_T$ plus some spurious ones. Then, $p_T$ also factorises
according to $\G_0$ and $\Delta = 0$. The Bayes factor still grows, but only
like $(k/2) \log n$, where $k$ is the number of additional parameters in $\G_0$
\citep{haughton88,chickering02}: the data overturn the prior only once $n$ is of
order $\exp\{2\beta E(\G_T, \G_0; w) / k\}$, exponential in $\beta$ rather
than linear.

We checked both results against the exhaustive enumerations of
Appendix~\ref{app:enumeration}. The bound of Lemma~\ref{theo:sidperturb} holds
for all prior DAGs at $N = 5, 6, 7$ and every value of $m$, and is attained at
$m = 1$ by the banded and the complete prior DAGs. The sparse prior DAGs are a
factor of about two below it, an arc reversal there rarely spoiling the
adjustment sets of both its endpoints. For Theorem~\ref{theo:minimax} and
Proposition~\ref{theo:minimax-upper}, we computed exactly the largest worst-case
$\Prob(\G_T)$ that any $\beta$ achieves, at $m = 1, 2, 3$ and $w = 0.5, 0.75,
1$: it falls inside the bracket in all $108$ configurations. At $w = 1$, the
guarantee is within a factor of $1.0$ to $3.9$ of the exact value and
$\beta^*(m)$ is larger than the exact maximiser by between $-3\%$ and $35\%$;
the upper bound is the looser of the two, between $0.3$ and $4.2$ orders of
magnitude above the exact value. At $w = 0.5$, the guarantee falls $1.0$ to
$8.7$ orders of magnitude below the exact value, at the rate the inflation
factor $\kappa_w$ predicts and because the optimum moves to a smaller $\beta$
at which the bracket on $Z(\beta, w)$ is itself wide; the upper bound, which
ignores SID altogether, loosens less, to $1.0$ to $6.8$ orders of magnitude
above it.

\section{Other Measures of Accuracy from Section~\ref{sec:sims}}
\label{app:kl}

We report the two measures of accuracy from Section~\ref{sec:sims} that LGE is
not a function of. The first is the divergence $\KL(\G, \G_T)$ from the
distribution of the reference BN (DAG $\G_T$) to that of the learned BN (DAG
$\G$), with the parameters of the latter fitted by maximum likelihood on the
same data the DAG was learned from. The second is the optimal adjustment
distance of \citet{henckel}, normalised by $N(N - 1)$ like $\wSID$; the
ancestor adjustment distance from the same paper is almost identical to it
across runs ($0.98$ correlation), so we do not report it. Table entries are
geometric means of the paired per-run ratios of each measure for LGE with $w =
0.5$ and for the baseline prior, as in Section~\ref{sec:sims}, by decay
$\beta$ and by $n / |\Theta|$, the sample size relative to the number of
parameters of the reference BN; values below $1$ mean that the DAG learned
with LGE is more accurate. Against the uniform prior, both measures favour LGE
in every cell of the tables by $\beta$ and $n/|\Theta|$, which we omit;
Tables~\ref{tab:kl-network} and~\ref{tab:aid-network} summarise them by
network.

\begin{table}[p]
\begin{center}
  \small
  \setlength{\tabcolsep}{4pt}
  \begin{tabular}{lrrrrrrcrrrrrr}
  \cline{2-7} \cline{9-14}
  & \multicolumn{6}{c}{correct $\G_0$} & &
    \multicolumn{6}{c}{$0.5|A_0|$ perturbations} \\
  \cline{2-7} \cline{9-14}
  $\beta$ & $1$ & $2$ & $5$ & $10$ & $20$ & $50$ & &
            $1$ & $2$ & $5$ & $10$ & $20$ & $50$ \\
  \cline{1-1} \cline{2-7} \cline{9-14}
  $2$   & $1.43$ & $1.17$ & $0.94$ & $0.97$ & $0.92$ & $0.91$ & &
          $1.46$ & $1.19$ & $0.96$ & $0.99$ & $0.96$ & $0.94$ \\
  $50$  & $1.31$ & $1.05$ & $0.86$ & $0.91$ & $0.86$ & $0.87$ & &
          $1.33$ & $1.09$ & $0.89$ & $0.94$ & $0.90$ & $0.91$ \\
  $200$ & $0.96$ & $0.79$ & $0.68$ & $0.75$ & $0.76$ & $0.79$ & &
          $1.04$ & $0.88$ & $0.77$ & $0.84$ & $0.83$ & $0.83$ \\
  $500$ & $0.65$ & $0.56$ & $0.53$ & $0.63$ & $0.67$ & $0.73$ & &
          $0.78$ & $0.69$ & $0.70$ & $0.80$ & $0.80$ & $0.83$ \\
  \cline{1-1} \cline{2-7} \cline{9-14}
  \end{tabular}
  \caption{$\KL(\G, \G_T)$ under LGE ($w = 0.5$), relative to the marginal
    uniform prior, by $\beta$ and $n/|\Theta|$, with a correct $\G_0$ (left)
    and with half of its arcs perturbed (right). Entries are geometric means of
    the paired per-run ratios over the four networks and $10$ replicates.}
  \label{tab:kl-decay}

  \vspace{1em}

  \begin{tabular}{cclrrrrcr}
  \cline{4-7} \cline{9-9}
  & & & \multicolumn{4}{c}{relative divergence} & & runs \\
  \cline{4-7} \cline{9-9}
  & & $\beta$ & ARTH150 & ECOLI70 & MAGIC-IRRI & MAGIC-NIAB & & LGE best \\
  \cline{3-3} \cline{4-7} \cline{9-9}
  \multirow{8}{*}{\rotatebox[origin=c]{90}{\footnotesize uniform}}
  & \multirow{4}{*}{\rotatebox[origin=c]{90}{\footnotesize correct}}
  & $2$   & $0.96$ & $0.88$ & $0.90$ & $0.91$ & & $0.83$ \\
  & & $50$  & $0.90$ & $0.81$ & $0.86$ & $0.82$ & & $0.98$ \\
  & & $200$ & $0.73$ & $0.68$ & $0.73$ & $0.62$ & & $0.99$ \\
  & & $500$ & $0.53$ & $0.60$ & $0.58$ & $0.50$ & & $1.00$ \\

  & \multirow{4}{*}{\rotatebox[origin=c]{90}{\footnotesize perturbed}}
  & $2$   & $0.97$ & $0.92$ & $0.93$ & $0.92$ & & $0.72$ \\
  & & $50$  & $0.91$ & $0.87$ & $0.88$ & $0.84$ & & $0.91$ \\
  & & $200$ & $0.75$ & $0.79$ & $0.78$ & $0.69$ & & $0.96$ \\
  & & $500$ & $0.58$ & $0.74$ & $0.70$ & $0.66$ & & $0.95$ \\
  \cline{3-3} \cline{4-7} \cline{9-9}
  \multirow{8}{*}{\rotatebox[origin=c]{90}{\footnotesize marginal uniform}}
  & \multirow{4}{*}{\rotatebox[origin=c]{90}{\footnotesize correct}}
  & $2$   & $1.70$ & $0.87$ & $0.86$ & $0.93$ & & $0.51$ \\
  & & $50$  & $1.58$ & $0.80$ & $0.81$ & $0.84$ & & $0.60$ \\
  & & $200$ & $1.29$ & $0.68$ & $0.69$ & $0.63$ & & $0.77$ \\
  & & $500$ & $0.93$ & $0.60$ & $0.55$ & $0.51$ & & $0.90$ \\

  & \multirow{4}{*}{\rotatebox[origin=c]{90}{\footnotesize perturbed}}
  & $2$   & $1.72$ & $0.91$ & $0.89$ & $0.94$ & & $0.48$ \\
  & & $50$  & $1.60$ & $0.87$ & $0.84$ & $0.86$ & & $0.58$ \\
  & & $200$ & $1.33$ & $0.79$ & $0.74$ & $0.71$ & & $0.69$ \\
  & & $500$ & $1.03$ & $0.73$ & $0.66$ & $0.68$ & & $0.81$ \\
  \cline{3-3} \cline{4-7} \cline{9-9}
  \end{tabular}
  \caption{$\KL(\G, \G_T)$ under LGE ($w = 0.5$), by network and pooled over
    the sample sizes, with a correct $\G_0$ and with half of its arcs
    perturbed. The outer row labels give the baseline prior; the last column
    is the fraction of runs in which LGE has the smaller $\KL(\G, \G_T)$.}
  \label{tab:kl-network}
\end{center}
\end{table}

\clearpage

\begin{table}[h]
\begin{center}
  \small
  \setlength{\tabcolsep}{4pt}
  \begin{tabular}{lrrrrrrcrrrrrr}
  \cline{2-7} \cline{9-14}
  & \multicolumn{6}{c}{correct $\G_0$} & &
    \multicolumn{6}{c}{$0.5|A_0|$ perturbations} \\
  \cline{2-7} \cline{9-14}
  $\beta$ & $1$ & $2$ & $5$ & $10$ & $20$ & $50$ & &
            $1$ & $2$ & $5$ & $10$ & $20$ & $50$ \\
  \cline{1-1} \cline{2-7} \cline{9-14}
  $2$   & $0.90$ & $0.72$ & $0.47$ & $0.35$ & $0.20$ & $0.08$ & &
          $1.21$ & $1.01$ & $0.76$ & $0.71$ & $0.62$ & $0.43$ \\
  $50$  & $0.75$ & $0.61$ & $0.35$ & $0.23$ & $0.13$ & $0.04$ & &
          $1.10$ & $0.94$ & $0.74$ & $0.72$ & $0.50$ & $0.40$ \\
  $200$ & $0.43$ & $0.33$ & $0.14$ & $0.08$ & $0.05$ & $0.02$ & &
          $0.91$ & $0.83$ & $0.65$ & $0.55$ & $0.43$ & $0.33$ \\
  $500$ & $0.07$ & $0.08$ & $0.03$ & $0.03$ & $0.02$ & $0.01$ & &
          $0.82$ & $0.64$ & $0.58$ & $0.62$ & $0.47$ & $0.41$ \\
  \cline{1-1} \cline{2-7} \cline{9-14}
  \end{tabular}
  \caption{Optimal adjustment distance under LGE ($w = 0.5$), relative to the
    marginal uniform prior, by $\beta$ and $n/|\Theta|$, with a correct $\G_0$
    (left) and with half of its arcs perturbed (right). Entries are geometric
    means of the paired per-run ratios over the four networks and $10$
    replicates.}
  \label{tab:aid-mu}

  \vspace{1em}

  \begin{tabular}{cclrrrrcr}
  \cline{4-7} \cline{9-9}
  & & & \multicolumn{4}{c}{relative optimal adjustment distance} & & runs \\
  \cline{4-7} \cline{9-9}
  & & $\beta$ & ARTH150 & ECOLI70 & MAGIC-IRRI & MAGIC-NIAB & & LGE best \\
  \cline{3-3} \cline{4-7} \cline{9-9}
  \multirow{8}{*}{\rotatebox[origin=c]{90}{\footnotesize uniform}}
  & \multirow{4}{*}{\rotatebox[origin=c]{90}{\footnotesize correct}}
  & $2$   & $0.33$ & $0.16$ & $0.36$ & $0.40$ & & $0.94$ \\
  & & $50$  & $0.28$ & $0.10$ & $0.28$ & $0.24$ & & $0.99$ \\
  & & $200$ & $0.17$ & $0.04$ & $0.19$ & $0.06$ & & $1.00$ \\
  & & $500$ & $0.07$ & $0.01$ & $0.09$ & $0.02$ & & $1.00$ \\

  & \multirow{4}{*}{\rotatebox[origin=c]{90}{\footnotesize perturbed}}
  & $2$   & $0.71$ & $0.64$ & $0.64$ & $0.61$ & & $0.82$ \\
  & & $50$  & $0.67$ & $0.62$ & $0.65$ & $0.47$ & & $0.87$ \\
  & & $200$ & $0.62$ & $0.53$ & $0.54$ & $0.36$ & & $0.93$ \\
  & & $500$ & $0.56$ & $0.52$ & $0.47$ & $0.44$ & & $0.93$ \\
  \cline{3-3} \cline{4-7} \cline{9-9}
  \multirow{8}{*}{\rotatebox[origin=c]{90}{\footnotesize marginal uniform}}
  & \multirow{4}{*}{\rotatebox[origin=c]{90}{\footnotesize correct}}
  & $2$   & $0.50$ & $0.17$ & $0.39$ & $0.44$ & & $0.86$ \\
  & & $50$  & $0.42$ & $0.11$ & $0.30$ & $0.26$ & & $0.94$ \\
  & & $200$ & $0.25$ & $0.04$ & $0.20$ & $0.07$ & & $1.00$ \\
  & & $500$ & $0.10$ & $0.01$ & $0.09$ & $0.02$ & & $1.00$ \\

  & \multirow{4}{*}{\rotatebox[origin=c]{90}{\footnotesize perturbed}}
  & $2$   & $1.06$ & $0.66$ & $0.68$ & $0.66$ & & $0.69$ \\
  & & $50$  & $1.00$ & $0.64$ & $0.69$ & $0.51$ & & $0.74$ \\
  & & $200$ & $0.92$ & $0.55$ & $0.57$ & $0.39$ & & $0.85$ \\
  & & $500$ & $0.84$ & $0.54$ & $0.50$ & $0.48$ & & $0.87$ \\
  \cline{3-3} \cline{4-7} \cline{9-9}
  \end{tabular}
  \caption{As Table~\ref{tab:aid-mu}, by network and pooled over the sample
    sizes. The outer row labels give the baseline prior.}
  \label{tab:aid-network}
\end{center}
\end{table}

\clearpage

\section{Single-Cell Data Experimental Setting}
\label{app:kidney}

This appendix provides the complete technical description of the analysis of the
proximal tubule single-cell data in Section~\ref{sec:kidney}.

\begin{figure}[p]
  \includegraphics[width=\linewidth]{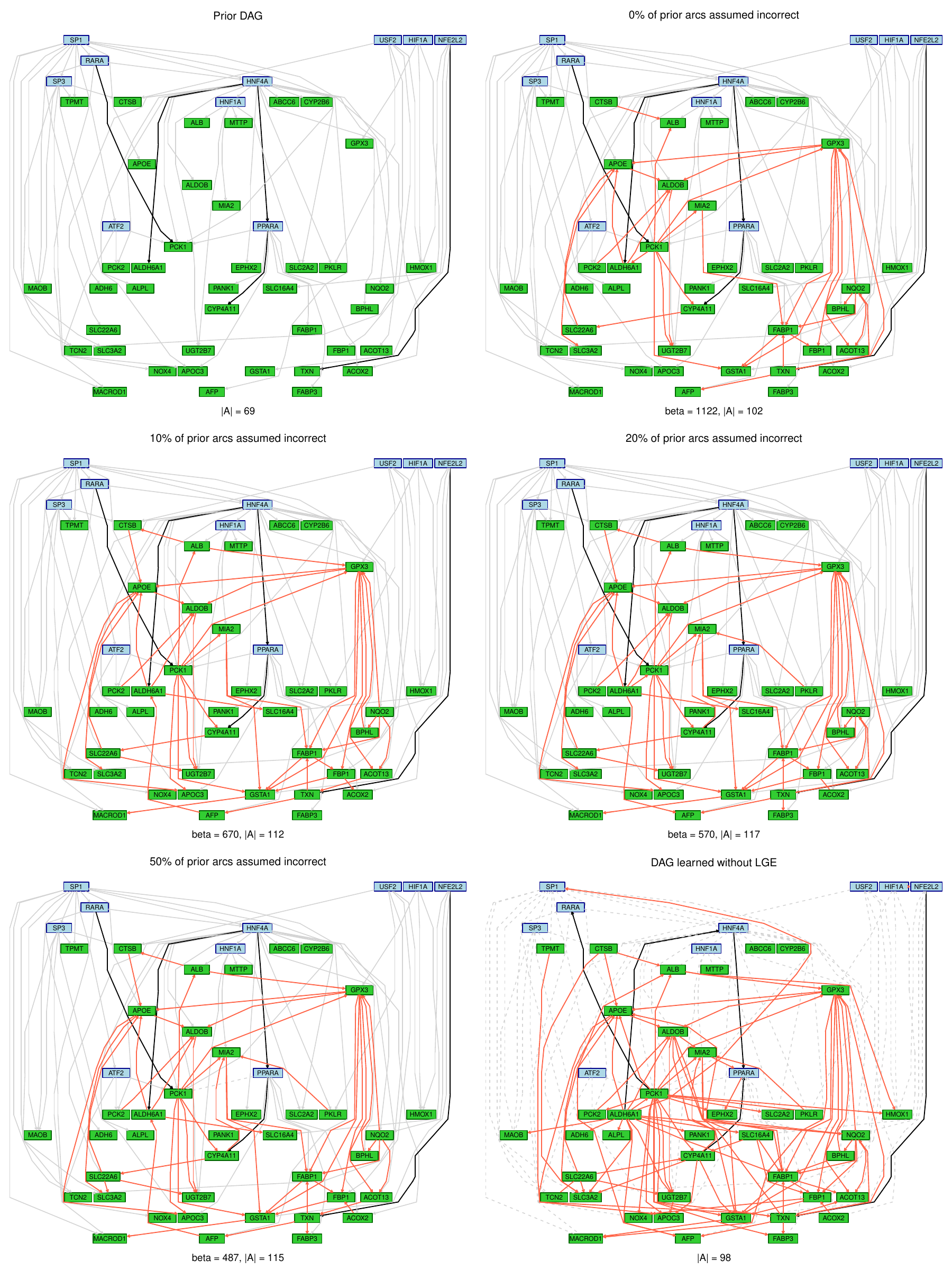}
  \caption{DAGs modelling the proximal tubule single-cell data: the prior DAG,
    the four posterior DAGs learned using LGE ($\beta = 1122, 670, 570, 487$,
    corresponding to $0\%$, $10\%$, $20\%$ and $50\%$ of the prior arcs assumed
    incorrect), and the DAG learned without LGE. Colours as in
    Figure~\ref{fig:sct}; the arcs of the prior DAG that a DAG lacks are
    dashed.}
  \label{fig:sct-all}
\end{figure}

\subsection{Data}
\label{app:kidney:data}

We used the counts and the cell metadata of the human kidney atlas of
\citet{lake23}, deposited in the Gene Expression Omnibus under accession number
GSE183276. We kept the $3087$ cells from healthy reference tissue (condition
``Ref'') annotated as proximal tubule cells (subclass ``PT'') in the
reference state (state ``reference''). We kept the counts of $50$ genes: $10$
transcription factors that regulate gene expression in the proximal tubule,
including HNF4A, HNF1A and PPARA, and $40$ genes that the Human Protein Atlas
\citep{hpa} reports as enriched in proximal tubular cells. Sequencing depth
enters the model as an additional node, defined for each cell as
$\log(1 + t)$, where $t$ is the total count of the cell over the $50$ genes.

\subsection{Prior DAG}
\label{app:kidney:prior}

We downloaded the human interactions in DoRothEA \citep{dorothea} from OmniPath
\citep{omnipath}. DoRothEA integrates four lines of evidence on the interactions
between transcription factors and their target genes: literature-curated
resources, ChIP-seq binding data, transcription factor binding sites predicted
on promoters, and regulons reverse-engineered from gene expression data. It
assigns each interaction a confidence level from A to E
\citep[Figure~5A]{dorothea}:
\begin{itemize}
  \item \emph{Level A (high confidence)}: interactions reported in at least
    one expert review or in at least two curated resources; those in a curated
    resource that also reports the sign of the regulation (activation or
    repression) and supported by other evidence; and those supported by all
    four lines of evidence.
  \item \emph{Level B (likely confidence)}: interactions supported by both
    curated resources and ChIP-seq data; and those supported by three lines of
    evidence, namely binding site predictions, inferred regulons, and either
    curated resources or ChIP-seq data.
  \item \emph{Levels C (medium confidence) and D (low confidence)}:
    interactions found in curated resources or ChIP-seq data, with and without
    supporting binding site predictions respectively.
  \item \emph{Level E}: interactions supported only by binding site
    predictions or by inferred regulons.
\end{itemize}
We retrieved only the interactions at
levels A and B, and kept those whose transcription factor and target are both
among the $50$ genes. When more than one interaction was reported for the same
pair of genes, we kept that with the highest confidence level, then a known
direction of regulation (activation or repression), and then the largest numbers
of supporting resources and references. After dropping self-loops, we removed a
minimum feedback arc set, computed with the \textbf{igraph} R package
\citep{igraph}, to make the resulting directed graph acyclic. Of the $71$
interactions retrieved, two pairs of genes (HNF4A and HNF1A, HNF4A and RARA) had
arcs in both directions; the feedback arc set removed one arc from each, HNF1A
$\rarr$ HNF4A and HNF4A $\rarr$ RARA, keeping the better-supported direction.
The resulting $\G_0$ has $69$ arcs, $65$ at level A and $4$ at level B.

\subsection{Structure Learning}
\label{app:kidney:learning}

We modelled each gene with a zero-inflated negative binomial distribution
given its parents \citep{zigdag}, and scored each DAG with the corresponding
BIC. The depth node is a parent of every gene and has no parents: it is included
in every candidate DAG, its arcs are never removed, and it is excluded from the
prior DAG and from the computation of SHD and SID in the energy
\mref{eq:energy}.

We learned each MAP DAG by tabu search, starting from the DAG containing only
the arcs from the depth node and moving by single-arc additions, deletions and
reversals between genes, for at most $4N = 200$ iterations. The search adds the
log-prior of LGE to the BIC of each candidate DAG, with $w = 0.83$ and $\beta
\in \{487, 570, 670, 1122\}$ (see Section~\ref{sec:kidney}). The data-driven DAG
uses the same approach with the BIC alone.

The model averaging follows \citet{aime11}. For each value of $\beta$, and for
the data-driven DAG, we drew $50$ subsamples of $2000$ of the $3087$ cells
without replacement, with the order of the genes shuffled to avoid any
dependence of the search on it, and learned one MAP DAG from each. The
strength of an arc is the fraction of these $50$ DAGs that contain it with the
same direction. The averaged DAG contains the arcs whose strength exceeds a
threshold that \citet{aime11} estimate from the empirical distribution of the
strengths. The estimated thresholds are $0.50$ for the data-driven DAG and
between $0.50$ and $0.54$ for the DAGs learned with LGE.

\subsection{Topological Orderings}
\label{app:kidney:orderings}

To compare the topological ordering of the genes in the DAGs of
Figure~\ref{fig:sct-all}, we computed their Spearman correlation, averaged over
$500$ random orderings of each DAG, since a DAG generally admits more than one.
The correlation between the orderings of $\G_0$ and of the data-driven DAG is
$0.15$. The orderings of the posterior DAGs move from $\G_0$ towards the
data-driven DAG as $\beta$ decreases: their correlation with that of $\G_0$
falls from $0.59$ at $\beta = 1122$ to $0.44$ at $\beta = 487$, while that with
the data-driven DAG rises from $0.15$ to $0.54$.

\clearpage

\renewcommand{\refname}{Supplementary References}

\input{supplement.bbl}
\end{supplement}